%% file: main.tex
\documentclass[11pt]{article}

\usepackage[preprint]{acl}

\usepackage{times}
\usepackage{latexsym}
\usepackage{amsthm,amsmath,amssymb}
\usepackage{mathrsfs}
\usepackage{booktabs}
\usepackage{algorithm}
\usepackage{multirow}
\usepackage{algorithmic}
\usepackage[switch]{lineno}

\usepackage{tcolorbox}
\newcommand{\model}{LAMO-3B}
\newcommand{\framework}{LAMO}
\usepackage{colortbl}
\definecolor{mycolor}{rgb}{0.925, 0.945, 0.980}
\definecolor{mygray}{rgb}{0.886, 0.886, 0.886}
\definecolor{mypink}{rgb}{0.992,0.921,0.929}
\definecolor{mygreen}{rgb}{0.901,0.950,0.937}
\usepackage[T1]{fontenc}

\usepackage[utf8]{inputenc}

\usepackage{microtype}

\usepackage{inconsolata}

\usepackage{graphicx}

\title{Towards Scalable Lightweight GUI Agents via Multi-role Orchestration}

\author{
 \textbf{Ziwei Wang\textsuperscript{1,2}},
 \textbf{Junjie Zheng\textsuperscript{1,2}},
 \textbf{Leyang Yang\textsuperscript{1,2}},
 \textbf{Sheng Zhou\textsuperscript{1 $\dagger$}},
 \textbf{Xiaoxuan Tang\textsuperscript{3}},\\
 \textbf{Zhouhua Fang\textsuperscript{3}},
 \textbf{Zhiwei Liu\textsuperscript{3}},
  \textbf{Dajun Chen\textsuperscript{3}},
   \textbf{Yong Li\textsuperscript{3 $\dagger$}},
    \textbf{Jiajun Bu\textsuperscript{1,2}}
\\
  \textsuperscript{1}Zhejiang Key Laboratory of Accessible Perception and Intelligent Systems, Zhejiang University\\
 \textsuperscript{2}College of Computer Science and Technology, Zhejiang University\\
 \textsuperscript{3}AntGroup
\\
\{wangziwei98, jjzheng0315, yangleyang, zhousheng\_zju, bjj\}@zju.edu.cn\\
\{tangxiaoxuan.txx, fangzhouhua.fzh, biao.lzw, chendajun.cdj, liyong.liy\}@antgroup.com
}

\begin{document}
\maketitle

\renewcommand\thefootnote{}
\footnotetext{$^{\dagger}$Co-corresponding authors.}
\footnotetext{Code: \href{https://github.com/BigTaige/LAMO}{https://github.com/BigTaige/LAMO}}
\renewcommand\thefootnote{\arabic{footnote}}

\begin{abstract}
Autonomous Graphical User Interface (GUI) agents powered by Multimodal Large Language Models (MLLMs) enable digital automation on end-user devices.
While scaling both parameters and data has yielded substantial gains, advanced methods still suffer from prohibitive deployment costs on resource-constrained devices. When facing complex in-the-wild scenarios, lightweight GUI agents are bottlenecked by limited capacity and poor task scalability under end-to-end episodic learning, impeding multi-agent systems (MAS) adaptation, while training multiple skill-specific experts remains costly.
\textit{Can we strike an effective trade-off in this cost–scalability dilemma, enabling lightweight MLLMs to participate in realistic GUI workflows?}
To address these challenges, we propose \framework\ framework, which endows a lightweight MLLM with GUI-specific knowledge and task scalability, allowing multi-role orchestration to expand their capability boundary for GUI automation. \framework\ combines role-oriented data synthesis with a two-stage training recipe: (i) supervised fine-tuning with Perplexity-Weighted Cross-Entropy optimization for knowledge distillation and visual perception enhancement, and (ii) reinforcement learning for role-oriented cooperative exploration. Via \framework, we develop a task-scalable native GUI agent \model\ supporting monolithic execution and MAS-style orchestration. When paired with advanced planners, as a plug-and-play policy executor, \model\ can continuously benefit from planner advances, enabling a higher performance ceiling. Extensive static and online evaluations validate the effectiveness of our designs.
\end{abstract}

\input{intro}

\input{related_work}

\input{method}
\input{experiments}
\input{conclusion}
\section{Acknowledgments}
This work was supported by the National Natural Science Foundation of China (Grant No.62372408).This work was supported by AntGroup Research Fund.

\section{Limitations}
We present a novel perspective for unlocking the potential of lightweight MLLMs in increasingly complex, in-the-wild scenarios via MAS adaptation. Despite the remarkable performance-to-size ratio of the proposed \model, several limitations remain and suggest directions for future work. First, owing to scaling law constraints, the limited parameter budget of \model\ poses a bottleneck for reasoning depth in complex GUI automation settings, especially for tasks involving long execution horizons (>$10$ steps). As such, achieving reliable performance in long-horizon GUI tasks still benefits from a hybrid approach that pairs lightweight \model\ with an advanced planner, which we argue is a promising paradigm. Although \model\ excels at grounding UI elements in mobile scenarios, its performance in desktop environments—where visual complexity is higher (e.g., spreadsheet scenarios and software-specific prior-dependent applications)—presents ongoing challenges (Figs.~\ref{fig:osworld-case1}, ~\ref{fig:osworld-case2}).
\bibliography{custom}

\appendix
\input{appendix}

\end{document}

%% file: intro.tex
\section{Introduction}\label{sec-intro}
The rapid evolution of Multimodal Large Language Models (MLLMs) has significantly propelled the development of agents for Graphical User Interfaces (GUIs)~\cite{gu2026towards}, marking a pivotal frontier in GUI automation~\cite{MobileAgentv3,agnet-s3}. These autonomous GUI agents are transforming how humans interact with digital systems by operating mobile/computer interfaces to accomplish user goals~\cite{survey2}.

GUI automation has progressed from static settings~\cite{androidcontrol} to increasingly complex in-the-wild online environments~\cite{rawles2024androidworld, xie2024osworld}. To address this challenging long-horizon reasoning task that integrates intent parsing, screen perception, history clues, and tool execution to achieve goals sequentially~\cite{hu2025memoryageaiagents}, current advanced methods have yielded substantial gains by scaling both parameters and data~\cite{qin2025ui,uivenus}. Scaling laws~\cite{scalinglaw} further endow large models with robust task scalability: they can build MAS via context engineering, alleviating the “lost-in-the-middle” issue~\cite{lostinthemiddle} and enabling efficient context management~\cite{MobileAgentv3,pgagent}, thus improving performance in navigating realistic GUI workflows. However, these gains come at a significantly higher system cost, making large-scale models impractical to deploy on resource-constrained devices.

Against this backdrop, lightweight GUI agents have drawn growing attention~\cite{wu2024atlas,r-vlm,hargui,uis1,lin2025showui}, with steady progress driven by post-training techniques such as supervised fine‑tuning (SFT) and reinforcement learning (RL).
Despite promising results on static, step-wise settings~\cite{androidcontrol}, small-scale MLLMs face two major constraints: inherent capacity bottlenecks due to their limited parameter size, and the end-to-end episodic learning framework~\cite{infigui-r1,guir1} couples high-level reasoning and low-level execution into a fixed pipeline, which suffers markedly when navigating realistic workflows~\cite{rawles2024androidworld,xie2024osworld}.
These constraints limit scalability and impede adaptation to MAS. Although training multiple skill-specific experts can mitigate these weaknesses, such methods remain costly~\cite{Co-EPG,MAPoRL}.
\textit{Can we strike an effective trade-off in this cost–scalability dilemma, enabling lightweight MLLMs to participate in realistic GUI workflows?}

To address these challenges, we propose \textbf{\framework}, a framework for \textbf{L}ightweight \textbf{A}gent \textbf{M}ulti-role \textbf{O}rchestration in GUI automation. \framework\ endows a lightweight MLLM with GUI-specific knowledge and task scalability through parameter sharing, enabling multi-role orchestration for MAS adaptation and expanding their capability boundary to solve increasingly complex in-the-wild scenarios. Unlike monolithic end‑to‑end agentic recipes~\cite{infigui-r1,lin2025showui,guir1}, \framework\ trains a lightweight MLLM to flexibly orchestrate skill-specific roles via role‑oriented data synthesis and a two‑stage training recipe: \textit{(i) SFT with a Perplexity‑Weighted Cross‑Entropy optimization for domain knowledge distillation, instruction following, and fine‑grained visual perception}; and \textit{(ii) Multi‑task RL for collaborative exploration and knowledge transfer across role‑oriented GUI tasks}.

Via \framework\ framework, we produce \textbf{\model}, a lightweight GUI agent that can flexibly orchestrate skill-specific roles to participate in realistic GUI workflows. In particular, \model\ functions as a reliable policy executor for precise low-level GUI execution; when paired with an advanced planner, as a plug-and-play policy executor, it can continually benefit from planner improvements, offering a higher performance ceiling than native monolithic models. Extensive experiments in both static (ScreenSpot-pro~\cite{screenspot-pro} and AndroidControl~\cite{androidcontrol}) and online (MiniWob++~\cite{miniwob}, AndroidWorld~\cite{rawles2024androidworld}, and OSWorld~\cite{xie2024osworld}) environments validate the effectiveness and potential of \framework.

Our main contributions are as follows:

\textbf{$\bullet$} We propose the \framework\ framework to endow a lightweight MLLM with task scalability for MAS adaptation, expanding its capability boundary to solve increasingly complex in-the-wild scenarios.

\textbf{$\bullet$} Via \framework, we train a task-scalable GUI agent, \model, that can be orchestrated into skill-specific roles for GUI-oriented tasks. Paired with advanced planners as the plug-and-play policy executor, \model's task scalability raises the performance ceiling for GUI automation.

\textbf{$\bullet$} Extensive experiments on both static and online benchmarks demonstrate the potential of \framework\ and the effectiveness of our designs.

%% file: related_work.tex
\section{Related Work}
Post-training techniques, such as SFT and RL, have advanced MLLM-powered GUI agents. MP-GUI~\cite{wang2025mp} enhances the GUI understanding of MLLMs via multi-perceiver augmentation, thereby improving its agentic performance. The powerful UI-TARS~\cite{qin2025ui}, trained via an SFT then RL strategy under a data flywheel, achieves milestone results in GUI automation. Furthermore, GUI-R1~\cite{guir1}, InfiGUI-R1~\cite{infigui-r1}, HAR-GUI~\cite{hargui}, and UI-S1~\cite{uis1} employ GRPO~\cite{shao2024deepseekmath} to further explore the potential of RL in GUI automation. However, despite strong static performance, these lightweight GUI agents degrade sharply online, widening the gap to realistic GUI workflows~\cite{probench}.
To address increasingly complex in-the-wild scenarios, MAS have emerged as a promising trend~\cite{hu2025memoryageaiagents,pgagent,zhang2025appagent}.
Leveraging the robust instruction-following of large-scale MLLMs, task scalability can be achieved via context engineering, which in turn enables MAS that orchestrate multiple skill-specific agents, exemplified by the Agent-S family~\cite{agents2,agnet-s3} and MobileAgent family~\cite{mobileagentv2,MobileAgentv3}, enable effective long-horizon reasoning.
However, current advanced MAS typically rely on large-scale MLLMs~\cite{Gemini2025,openai2025gpt5}, which are suboptimal for low-level GUI execution and thus require specialized, large-parameter GUI experts for reliable actuation, for example, Agent-S2~\cite{agents2} deploy multiple large-scale, GUI-specific executors, including UI-TARS-72B-DPO~\cite{qin2025ui} for visual grounding, Tesseract OCR~\cite{tesseract2025} for textual grounding, and UNO~\cite{uno} for structural grounding, resulting in prohibitive system cost. Meanwhile, lightweight GUI agents trained via end-to-end episodic learning endow poor task scalability, restricting their adaption to MAS workflows. This has driven demand for task-scalable lightweight GUI agents.

%% file: method.tex
\begin{figure*}
    \centering
    \includegraphics[width=1\linewidth]{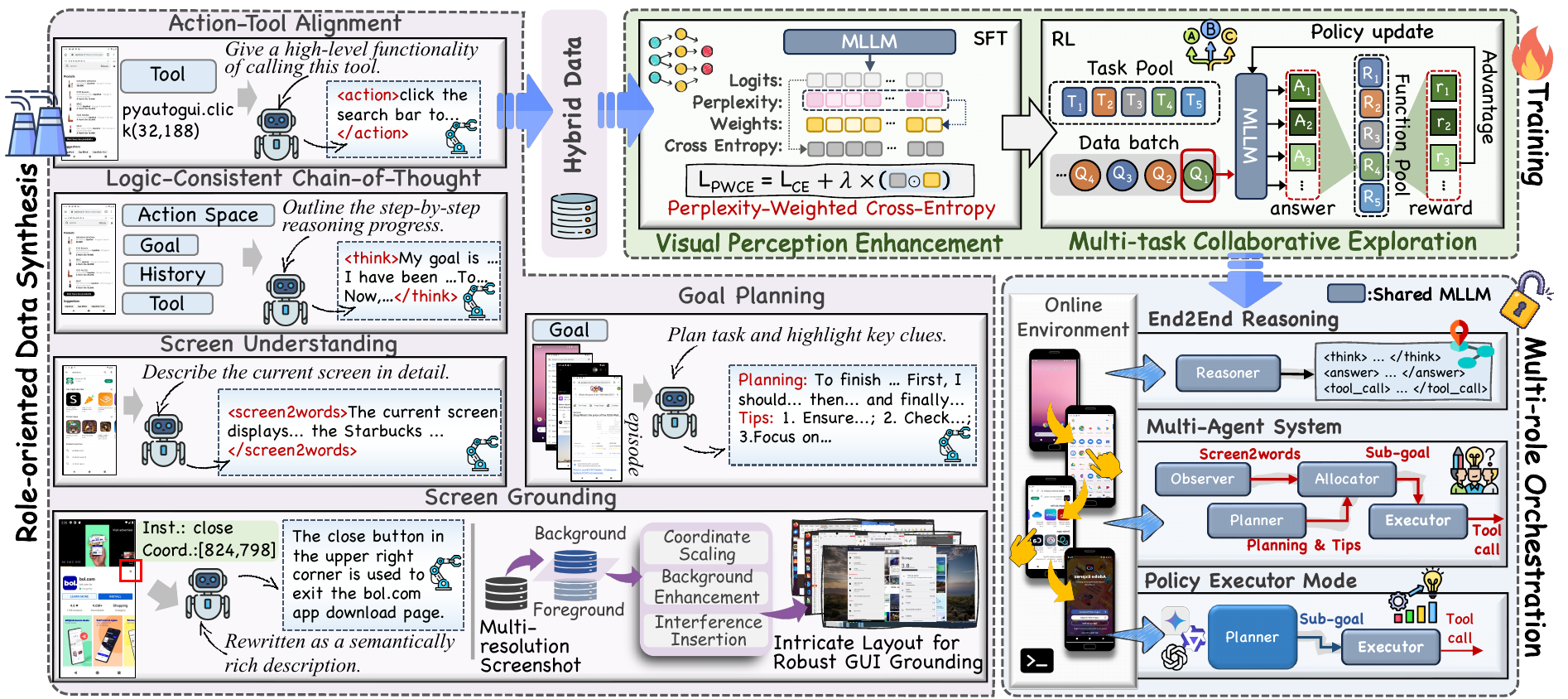}
    \caption{Overview of the Lightweight Agent Multi‑role Orchestration (\framework) framework. \framework\ integrates role-oriented data synthesis with a two-round training recipe to enhance screen perception, long-horizon reasoning, and multi-role orchestration. It enables versatile inference modes, allowing a lightweight MLLM to function as end-to-end monolithic agent, coordinated MAS, or plug-and-play executor paired with advanced planners. Such scalability expands the capability boundary of lightweight MLLMs in GUI automation via MAS adaptation.}
    \label{fig:framework}
\end{figure*}

\section{Problem Formulation}
For step-wise GUI tasks (e.g., grounding and screen QA), the MLLM $\mathcal M_{\theta}(\cdot)$ performs end-to-end inference as $y_{k}$=$\mathcal M_{\theta}(y_{<k}\mid \mathcal I, o)$, where $\mathcal I$ denotes the instruction and $o$ is a screenshot image. For long-horizon GUI tasks, let $\mathcal{T}$ denote an episode with an overall goal $\mathcal{G}$, where $\mathcal{T} = (\mathcal{G}, (o_1, {a}_1), ..., (o_n, a_n))$ and each observation $o_t$ is the screenshot at $t$-th step. The atomic action $a_t \in \mathcal{A}$ is an operation executed by the agent, with $\mathcal{A}$ denoting the predefined {\tt PyAutoGUI}-style action space. The agentic task is formulated as a Markov Decision Process $P({a}_t \mid \mathcal{G}, \mathcal I, o_{\leq t}, {a}_{<t})$ that drives the agent step by step toward achieving the goal. 

\section{Methodology}
Fig.~\ref{fig:framework} overviews the \textbf{\framework} framework and the key designs are detailed below.

\subsection{Role-oriented Data Synthesis}\label{sec-role-oriented-data}
Data-centric native GUI agents, powered by post-training techniques such as SFT and RL, show significant promise~\cite{qin2025ui,hargui}. We observe that lightweight MLLMs, though weak on long-horizon tasks where the agent must handle screen analysis, policy decisions, and tool invocation simultaneously, perform reliably when these components are handled independently.
Following this insight, we aim to decompose high-level reasoning into a GUI-oriented sub-task flow: \textit{(i) progressively improving sub-task performance to achieve overall gains, and (ii) using parameter sharing and context engineering to orchestrate the model into skill-specific roles that communicate and collaborate efficiently for GUI automation}.

To achieve this goal, we introduce a role‑oriented data synthesis strategy that decomposes GUI automation into five core capabilities: \textbf{\textit{Action–Tool Alignment (ATA)}}\footnote{This data format also supports tool-call summarization, improving history reconstruction at inference.} for mapping high-level instructions to low-level executable tools; \textbf{\textit{Logic‑Consistent CoT (LCC)}} for analyzing in‑context clues and yielding logically coherent reasoning; \textbf{\textit{Screen Understanding (SU)}} for interpreting screen functionality and screen details; \textbf{\textit{Goal Planning (GP)}} for decomposing overall goals into executable subtasks and identifying the key considerations for accomplishing these tasks; and \textbf{\textit{Screen Grounding (SG)}} for fine‑grained spatial and UI layout perception. We synthesize skill‑specific training data for each category using teacher models: Qwen‑2.5‑VL‑72B‑Instruct~\cite{bai2025qwen2.5} ($\mathcal M_1^{\mathbb T}$) for ATA and SG, and Gemini‑2.5‑Pro~\cite{Gemini2025} ($\mathcal M_2^{\mathbb T}$) for SU, LCC, and GP.

The ATA task trains the agent as a policy executor by synthesizing an action-aligned description
$\mathcal{C}_{\text{act}}=\mathcal{M}_1^{\mathbb{T}}(\mathcal{I}_{\text{Tool}},\, o_k,\, a_k)$,
where $\mathcal{C}_{\text{act}}$ verbalizes the atomic action $a_k$.
The LCC task equips the agent with long-horizon reasoning by synthesizing step-wise logically rigorous CoT 
$\mathcal{C}_{\text{CoT}}=\mathcal{M}_2^{\mathbb{T}}(\mathcal{I}_{\text{CoT}},\, \mathcal{G},\, o_k,\, a_{\le k})$,
where $\mathcal{C}_{\text{CoT}}$ provides the rationale at step $k$.
The SU task trains the agent as a screen observer by synthesizing multi-grained screen descriptions
$\mathcal{C}_{\text{S2W}}=\mathcal{M}_2^{\mathbb{T}}(\mathcal{I}_{\text{S2W}},\, o_k)$,
where $\mathcal{C}_{\text{S2W}}$ summarizes screen functionality, layout, and key UI elements.
The GP task trains the agent as a planner. We synthesize planning supervision as
$(\mathcal C_{\text{Plan}},\, \mathcal C_{\text{Tips}}) = \mathcal M_2^{\mathbb T}(\mathcal I_{\text{Plan}},\, \mathcal G,\, o_{\leq k},\, a_{\leq k})$,
where $\mathcal C_{\text{Plan}}$ describes subtasks and $\mathcal C_{\text{Tips}}$ provides key considerations for accomplishing $\mathcal G$.
The instructions $\mathcal{I}_{\text{Tool}}$, $\mathcal{I}_{\text{CoT}}$, $\mathcal{I}_{\text{S2W}}$, and $\mathcal I_{\text{Plan}}$
are task-specific prompts for ATA, LCC, SU, and GP, respectively.

For SG, we target two practical challenges faced by GUI agents: \textbf{\textit{(i)}} \textit{limited semantic understanding of element descriptions, especially for semantically sparse elements that are common in real scenarios;} \textbf{\textit{(ii)}} \textit{difficulty grounding targets in complex layouts with abundant distracting signals.}
For the first challenge, we enrich the original element description $\mathcal{C}_{\text{orig}}$ into a semantically rich caption:$\mathcal{C}_{\text{rich}}=\mathcal{M}_1^{\mathbb{T}}(\mathcal{I}_{G},\, o_k,\, \mathcal{C}_{\text{orig}})$.
We then distill $\mathcal{C}_{\text{rich}}$ together with the element coordinates $\mathcal{P}_{\text{point}}$ into the agent $\mathcal{M}_{\theta}$ by training it to predict $(\mathcal{C}_{\text{rich}},\, \mathcal{P}_{\text{point}})$
given $(o_k,\, \mathcal{C}_{\text{orig}})$ under the training prompt $\mathcal{I}^{*}_{G}$, fostering fine-grained semantic understanding of UI elements. $\mathcal{I}_{G}$ and $\mathcal{I}^{*}_{G}$ denote the prompts used for data synthesis and for training/inference, respectively.

For the second challenge, we perform rule‑based augmentation on the grounding metadata $\mathcal D$: we sample foregrounds $\mathcal D^{+}$ and backgrounds $\mathcal D^{-}$ from $\mathcal D$, take a screen from $\mathcal D^{-}$ as the background view $\mathcal O_{\text{back}}$, and overlay a target $(o_i, P_{\text{point}}^i)$ and multiple distractor screens from $\mathcal D^{+}$ with random scaling, yielding a cluttered intricate‑layout screen $\ddot{\mathcal O}$ with scaled coordinates $\ddot P_{\text{point}}^i$. In this way, each meta sample $(o_i, P_{\text{point}}^i, \mathcal C_{\text{orig}}) \in \mathcal D$ is converted into an Intricate‑Layout Grounding (ILG) sample $(\ddot{\mathcal O}, \ddot P_{\text{point}}^i, \mathcal C_{\text{orig}})$. This rule‑based procedure ensures controllable augmentation quality and enables low‑cost synthesis of large‑scale ILG data to improve grounding robustness in complex real‑world screen states (Tab.~\ref{ablation}). Instructions and ILG data synthesis details are in App.~\ref{appix-instruction} and~\ref{appix-ILG}.

\subsection{Visual Perception Enhancement}
SFT is a reasonable solution to equip MLLMs with domain knowledge during post-training \cite{wu2024atlas}, but its effectiveness degrades on agentic tasks requiring fine‑grained visual perception, especially UI grounding with accurate numerical coordinates. Our empirical evaluations find that while SFT improves textual learning, yet predicted coordinates exhibit small but systematic deviations from the ground truth, dicating limited spatial awareness. Qualitative analysis suggests that coordinate tokens often exhibit higher perplexity than textual tokens during SFT, reflecting the inherent uncertainty of numerical prediction.

To mitigate this, we introduce the {Perplexity-Weighted Cross-Entropy (PWCE)} loss function, which reweights tokens according to their perplexity: high‑perplexity tokens, particularly coordinates, receive larger loss weights, steering optimization toward uncertain, spatially critical outputs and enhancing screen details perception.

Specifically, let the shifted last-layer hidden states of the LLM for next-token prediction be
$\tilde{h} \in \mathbb{R}^{b \times l \times d}$ and the corresponding labels be
$\tilde{y} \in \mathbb{R}^{b \times l}$, where $b$, $l$, and $d$ denote batch size,
sequence length, and hidden dimension, respectively.
The model produces logits $h^* = \tilde{h} W^\top \in \mathbb{R}^{b \times l \times V}$, where $W$ is the embedding matrix and $V$ is the vocabulary size. The standard cross-entropy loss over the sequence is $\mathcal{L}_{\text{CE}} = \mathrm{CE}(h^*, \tilde{y})$.
We construct a binary mask $M$ to indicate tokens generated after the input.
For each masked index $i \in M$, we compute probabilities
$p_i = \mathrm{softmax}(h^*_i)$, token entropy
$E_i = -\sum_{v=1}^V p_{i,v} \log(p_{i,v} + \epsilon)$, and perplexity
$\mathrm{PPL}_i = \min\big(\exp(\sqrt{E_i}),\, \beta\big)$, where $\epsilon$ and
$\beta$ are hyperparameters. Let $\overline{\mathrm{PPL}}$ denote the mean perplexity
over tokens in $M$. The perplexity-oriented weights and the perplexity-weighted loss can be formulated as,
\begin{equation}
w_i =
\frac{
    1 + \alpha \,\dfrac{\mathrm{PPL}_i}{\overline{\mathrm{PPL}} + \epsilon}
}{
    \dfrac{1}{|M|} \sum_{j \in M}
    \left(
        1 + \alpha \,\dfrac{\mathrm{PPL}_j}{\overline{\mathrm{PPL}} + \epsilon}
    \right)
},
\end{equation}
\begin{equation}
\mathcal{L}_{\text{PW}} = \frac{1}{|M|} \sum_{i \in M} w_i \cdot \mathrm{CE}(h^*_i, \tilde{y}_i).
\end{equation}

Finally, the PWCE objective is defined as $\mathcal{L}_{\text{PWCE}} =  \mathcal{L}_{\text{CE}} + \lambda \,\mathcal{L}_{\text{PW}}$, where $\alpha$ and $\lambda$ are hyperparameters. PWCE dynamically assigns higher weights to numerical coordinate tokens and key contextual tokens with higher generation perplexity, guiding the model to percept screen details.

\subsection{Multi-task Collaboration Exploration}\label{sec:MAS_RL}
Following SFT, the model acquires extensive GUI-specific knowledge, improves instruction following, and adapts to role orchestration. Then, we perform a second round of RL with multi-task cooperative exploration to facilitate the discovery of optimal reasoning pathways in role-oriented tasks\footnote{Our empirical evaluations reveal that multi‑task RL facilitates the acquisition of shared representations and inter‑task dependencies across GUI‑related tasks, enabling effective knowledge transfer and improving overall performance.}. 

Specifically, we curate the hybrid data (ATA, SU, GP, SG, and the formatted step‑wise agentic task LCC) to build a task pool\footnote{Each task is assigned a unique tag for rule-based hybrid reward computation.} and construct a function pool that stores task‑specific, rule‑based reward functions for advantage estimation.
For SU and GP, the reward is defined as the normalized similarity between prediction and label under the TF‑IDF metric,
$r_{\mathrm{agent}} = \mathrm{norm}(\mathrm{TF\text{-}IDF}(y_{\text{pred}}, y_{\text{label}})) \in [0, 1]$.
For SG, we extract coordinate points from the predictions and, following~\citealp{hargui}, compute their geometric distance to the ground truth to define the reward $r_{\mathrm{agent}}$.
For ATA and agentic tasks, exemplified by \texttt{pyautogui.write(message='\$50')}, we parse the tool class (\texttt{write}) and tool value (\texttt{'\$50'}) from the prediction, compute binary $(0/1)$ scores $r_{\text{class}}$ and $r_{\text{val}}$ via string matching, and aggregate them as $r_{\text{agent}} = r_{\text{class}} + r_{\text{val}}$. For coordinate-based tool values (e.g., \texttt{click}, \texttt{swipe}, \texttt{dragTo}), we reuse the SG reward function.  
A length-penalty $r_{\text{penalty}} = -\varphi \cdot \frac{\mathrm{length}(y_{pred})}{L_{\text{max}}}$ is added to all reward functions to avoid uncontrolled output length.
Finally, we employ GRPO~\cite{shao2024deepseekmath} to conduct multi-task cooperative exploration.
\begin{table*}[ht!]\scriptsize
  \renewcommand{\arraystretch}{1} 
  \setlength{\tabcolsep}{0.42mm}
  \centering
  \begin{tabular}{l *{7}{ccc} ccc}
\toprule
\multirow{2}{*}{\textbf{Method}} & \multicolumn{3}{c}{\textbf{Development}} 
             & \multicolumn{3}{c}{\textbf{Creative}} 
             & \multicolumn{3}{c}{\textbf{CAD}} 
             & \multicolumn{3}{c}{\textbf{Scientific}} 
             & \multicolumn{3}{c}{\textbf{Office}} 
             & \multicolumn{3}{c}{\textbf{MacOS}} 
             & \multicolumn{3}{c}{\textbf{Avg.}} \\
\cmidrule(lr){2-4} \cmidrule(lr){5-7} \cmidrule(lr){8-10} \cmidrule(lr){11-13} 
\cmidrule(lr){14-16} \cmidrule(lr){17-19} \cmidrule(lr){20-22}
 & Text & Icon & Avg.
 & Text & Icon & Avg.
 & Text & Icon & Avg.
 & Text & Icon & Avg.
 & Text & Icon & Avg.
 & Text & Icon & Avg.
 & Text & Icon & Avg. \\
\midrule
\rowcolor{mygray}
\multicolumn{22}{l}{\textit{Generic Models}}\\
Claude Computer Use\cite{claude} & 22.0 & 3.9 & 12.6 & 25.9 & 3.4 & 16.8 & 14.5 & 3.7 & 11.9 & 33.9 & 15.8 & 25.8 & 30.1 & 16.3 & 26.9 & 11.0 & 4.5 & 8.1 & 23.4 & 7.1 & \cellcolor{mycolor}17.1 \\
Qwen2.5-VL-3B\cite{bai2025qwen2.5} & 38.3 & 3.4 & 21.4 & 40.9 & 4.9 & 25.8 & 22.3 & 6.3 & 18.4 & 44.4 & 10.0 & 29.5 & 48.0 & 17.0 & 40.9 & 33.6 & 4.5 & 20.4 & 37.8 & 6.6 &\cellcolor{mycolor}25.9 \\
Qwen2.5-VL-7B\cite{bai2025qwen2.5} & 51.9 &4.8 &29.1& 36.9& 8.4& 24.9& 17.8& 1.6& 13.8& 48.6& 8.2& 31.1 &53.7& 18.9& 45.7& 34.6 &7.9& 22.4& 39.9 &7.6 &\cellcolor{mycolor}26.8 \\
Kimi-VL-A3B-Instruct\cite{team2025kimi}
& -- & -- & -- 
& -- & -- & -- 
& -- & -- & --
& -- & -- & --
& -- & -- & -- 
& -- & -- & -- 
& -- & -- & \cellcolor{mycolor}34.5 \\
\midrule
\rowcolor{mygray}
\multicolumn{22}{l}{\textit{GUI-specific Models}}\\
UGround-72B\cite{uground} &  \underline{55.8} & 4.8 & 31.1 & \textbf{54.0} & \textbf{10.5} & \textbf{35.8} & 16.8  & 4.7	& 13.8 & \textbf{70.8} & 22.7 & \textbf{50.0} & 61.0 & 18.9 & 51.3 & 40.2 & 7.9 & 25.5 & -- & -- & \cellcolor{mycolor}34.5\\
SeeClick\cite{cheng2024seeclick} & 0.6 & 0.0 & 0.3 & 1.0 & 0.0 & 0.6 & 2.5 & 0.0 & 1.9 & 3.5 & 0.0 & 2.0 & 1.1 & 0.0 & 0.9 & 2.8 & 0.0 & 1.5 & 1.8 & 0.0 & \cellcolor{mycolor}1.1 \\
UGround-7B\cite{uground} & 26.6 & 2.1 & 14.7 & 27.3 & 2.8 & 17.0 & 14.2 & 1.6 & 11.1 & 31.9 & 2.7 & 19.3 & 31.6 & 11.3 & 27.0 & 17.8 & 0.0 & 9.7  & 25.0 & 2.8 & \cellcolor{mycolor}16.5 \\
OS-Atlas-7B\cite{wu2024atlas}	& 33.1 & 1.4& 17.7 & 28.8 & 2.8 & 17.9 & 12.2 &	4.7 & 10.3 & 37.5 & 7.3 & 24.4 & 33.9 &	5.7 & 27.4 & 27.1 &	4.5 & 16.8 & 28.1 &	4.0 & \cellcolor{mycolor}18.9 \\
GUI-R1-7B\cite{guir1} & 49.4 & 4.8 & -- & 38.9 & 8.4 & -- & 23.9 & 6.3 & -- & 55.6 & 11.8 & -- & 58.7 & \underline{26.4} & -- & \underline{42.1} & \underline{16.9} & -- & -- & -- & \cellcolor{mycolor}-- \\
UI-S1-7B\cite{uis1} & -- & -- & -- 
& -- & -- & -- 
& -- & -- & --
& -- & -- & --
& -- & -- & -- 
& -- & -- & -- 
& -- & -- & \cellcolor{mycolor}30.6 \\
UI-TARS-7B\cite{qin2025ui} & \textbf{58.4} & \textbf{12.4} & \textbf{36.1} & \underline{50.0} & \underline{9.1} & \underline{32.8} & 20.8 & 9.4 & 18.0 & \underline{63.9} & \textbf{31.8} & \textbf{50.0} & 63.3 & 20.8 & 53.3 & 30.8 & \underline{16.9} & 24.5 & 47.8 & \underline{16.2} & \cellcolor{mycolor}\underline{35.7} \\
OS-Atlas-4B\cite{wu2024atlas} & 7.1 & 0.0 & 3.7 & 3.0 & 1.4 & 2.3 & 2.0 & 0.0 & 1.5 & 9.0 & 5.5 & 7.5 & 5.1 & 3.8 & 4.8 & 5.6 & 0.0 & 3.1 & 5.0 & 1.7 & \cellcolor{mycolor}3.7\\
GUI-R1-3B\cite{guir1} & 33.8 & 4.8 & -- & 40.9 & 5.6 & -- & 26.4 & 7.8 & -- & {61.8} & 17.3 & -- & 53.6 & 17.0 & -- & 28.1 & 5.6 & -- & -- & -- & \cellcolor{mycolor}-- \\
UI-TARS-2B\cite{qin2025ui} & 47.4 & 4.1 & 26.4 & 42.9 & 6.3 & 27.6 & 17.8 & 4.7 & 14.6 & 56.9 & 17.3 & 39.8 & 50.3 & 17.0 & 42.6 & 21.5 & 5.6 & 14.3 & 39.6 & 8.4 & \cellcolor{mycolor}27.7 \\
InfiGUI-R1-3B\cite{infigui-r1}& 51.3 & \textbf{12.4} & \underline{32.4} & {44.9} & 7.0 & {29.0} & \underline{33.0} & \underline{14.1} & \underline{28.4} & 58.3 & 20.0 & \underline{41.7} & \textbf{65.5} & \textbf{28.3} & \textbf{57.0} & \textbf{43.9} & 12.4 & \textbf{29.6} & \textbf{49.1} & 14.1 & \cellcolor{mycolor}\underline{35.7} \\
\midrule
\model\ (ours) & 52.6 & \underline{9.7} & 31.8 & 40.9 & \underline{9.1} & 27.6 & \textbf{39.1} & \textbf{21.9} & \textbf{34.9} & 52.8 & \underline{27.3} & \underline{41.7} & \underline{65.0} & 24.5 & \underline{55.7} & 35.5 & \textbf{21.3} & \underline{29.1} & \underline{47.9} & \textbf{17.1} & \cellcolor{mycolor}\textbf{36.1} \\
\bottomrule
\end{tabular}
\caption{Grounding accuracy on ScreenSpot-pro. \textbf{Bold} represents the best results, \underline{underlined} is the second best.}
\label{screenspot-pro}
\end{table*}
\subsection{Multi-role Orchestration}\label{sec-mas}
Using the \framework\ framework, we yield a task-scalable GUI agent \model\ ($\mathcal M_{\Theta}$), facilitating the following inference modes for MAS adaptation:

\noindent\textbf{End-to-End Reasoning.}
\model\ can serve as a reasoner with a ReAct-style\cite{react} agentic reasoning paradigm, enabling it to attend to details in the dynamic context (e.g., clues from historical interactions and tool descriptions) and to analyze the current screen state to make decisions with structured outputs.
At time step $t$, given observations $o_{\le t}$ and interaction history $a_{<t}$, \model\ jointly encodes visual and textual inputs and outputs a structured decision:
\begin{equation}
\mathcal{S}_t = \mathcal{M}_{\Theta}\!\left(\mathcal{G},\, \mathcal{I}_{\mathrm{e2e}},\, o_{\le t},\, a_{<t}\right).
\end{equation}
The instruction $\mathcal{I}_{\mathrm{e2e}}$ guides $\mathcal{M}_{\Theta}$ to perform step-wise reasoning and generate structured $\mathcal{S}_t$, formatted as
\texttt{<think>CoT</think>}
\texttt{<answer> action decision </answer>}
\texttt{<tool\_call> atomic action $a_t$</tool\_call>}.

\begin{algorithm}[t]
\small
\caption{MAS workflow built on \model}
\label{agm-mas}
\begin{algorithmic}[1]

\STATE \textbf{Initialization:} overall goal $\mathcal{G}$, 
       time step $t$, 
       screenshot $o_t$, 
       GUI agent $\mathcal{M}_\Theta$, 
       action space $\mathcal{A}$, 
       max execution steps $T_{\mathrm{max}}$, instructions [$\tilde{\mathcal{I}}_{\mathrm{obs}}$, $\tilde{\mathcal{I}}_{\mathrm{plan}}$, 
       $\tilde{\mathcal{I}}_{\mathrm{act}}$, $\tilde{\mathcal{I}}_{\mathrm{exec}}$].

\REPEAT
    \STATE \textbf{Observation:} 
           $\mathcal C_{\mathrm{s2w}} \gets \mathcal{M}_\Theta^{\mathrm{Observer}}(\tilde{\mathcal{I}}_{\mathrm{obs}}, o_t)$
           \textit{// Provide richly detailed semantic descriptions of the screen.}

    \STATE \textbf{Planning:} 
           $(\mathcal C_{\mathrm{plan}}, \mathcal C_{\mathrm{tips}}) \gets \mathcal{M}_\Theta^{\mathrm{Planner}}(\tilde{\mathcal{I}}_{\mathrm{plan}}, \mathcal{G}, o_{\leq t})$
           \textit{// Interpret the goal, decompose it into subtasks, and provide actionable guidelines during execution.}

    \STATE \textbf{Allocation:} 
           $\mathcal C_{\mathrm{action}} \gets \mathcal{M}_\Theta^{\mathrm{Allocator}}(\tilde{\mathcal{I}}_{\mathrm{act}}, o_{t},a_{<t},$ $\mathcal C_{\mathrm{s2w}}, \mathcal C_{\mathrm{plan}}, \mathcal C_{\mathrm{tips}})$
           \textit{// Assign a single executable action for the current screen based on historical interactions and contextual clues.}

    \STATE \textbf{Execution:} 
           $a_t \gets \mathcal{M}_\Theta^{\mathrm{Executor}}(\tilde{\mathcal{I}}_{\mathrm{exec}}, o_t, \mathcal  C_{\mathrm{action}})$
           \textit{// Based on the instruction, select the optimal tool from the action space and execute it within the environment.}

\UNTIL{$\mathcal{G}$ reached or $t > T_{\mathrm{max}}$}
\end{algorithmic}
\end{algorithm}

\noindent\textbf{Multi-Agent System (MAS).}
To address the limited reasoning of lightweight MLLMs under the Scaling Laws~\cite{scalinglaw} and the "lost-in-the-middle" issue in long-horizon interactions~\cite{pgagent}, we orchestrate \model\ into a parameter-shared MAS that decomposes GUI automation into skill-specific roles. As shown in Fig.~\ref{fig:framework} and Alg.~\ref{agm-mas}, \model\ uses a shared backbone to instantiate four agents:  Observer $\mathcal M_{\Theta}^{Observer}$, Planner $\mathcal M_{\Theta}^{Planner}$, Allocator $\mathcal M_{\Theta}^{Allocator}$ and Executor $\mathcal M_{\Theta}^{Executor}$, which jointly reduce task complexity, improve context management, and enable low-hallucination reasoning.

\noindent\textbf{Policy Executor Mode.}
Given the increasing complexity and long-horizon nature of GUI automation (e.g., computer-use and cross-APP scenarios), success often depends on implicit GUI-specific priors and strong planning capabilities in the underlying foundation MLLMs~\cite{agents2}. Yet, constrained by limited parameters, lightweight GUI agents struggle to align with real-world environments~\cite{probench}.
To bridge this gap, \framework\ enables \model\ to act as a plug-and-play policy executor to reliably interacts with the environment, worked with an advanced large-scale MLLM planner (e.g., Gemini-2.5-Pro or GPT-5) to drive GUI automation. Compared with GUI agents with limited task scalability~\cite{infigui-r1}, {\model\ can evolve alongside advanced planners, yielding a higher performance ceiling}.

Specifically, the execution can be expressed as follows, an advanced MLLMs act as a planner to provide an executable high-level instruction, then \model\ converts it into atomic screen actions. The process can be formulated as,
\begin{equation}
    \mathcal C^*_{action} = \mathcal M^{Planner}_{\Delta}(\mathcal I^*_{\Delta}, \mathcal G, o_{\leq t}, a_{<t}),
\end{equation}
\begin{equation}
    a_t = \mathcal M_{\Theta}^{Executor}(\tilde{\mathcal{I}}_{\mathrm{exec}}, o_{t}, \mathcal C^*_{action}),
\end{equation}

where $\mathcal M_{\Delta}^{Executor}$ denotes an advanced foundational MLLM, conditioned on the instruction $\mathcal I^*_{\Delta}$. Our multi-role orchestration strategy enables \model\ to adapt seamlessly to both monolithic reasoning and cooperative multi-agent execution, thereby enhancing robustness and scalability in GUI automation. Case study in App.~\ref{appix-case}.

%% file: experiments.tex
\section{Experiments}\label{sec-experiments}

\subsection{Implementation Details}
Using the constructed hybrid dataset (Sec.~\ref{sec-role-oriented-data}; see App.~\ref{appix-datacuration} for data curation details) and our training recipe, we develop \model\ from Qwen2.5-VL-3B-Instruct~\cite{bai2025qwen2.5} under the \framework\ framework. The data distillation stage applies SFT for 1 epoch with a learning rate of 4e-6, warmup ratio 0.03, global batch size 256, and LoRA (rank 128, alpha 256, dropout 0.001). In the RL stage, the vision backbone is frozen while the merge layer and LLM are trained with GRPO for 1 epochs at learning rate 1e-6, rollout batch size 32, generating 8 rollouts per sample. For hparams, we set $\epsilon$=1e-12, $\beta$=1.5, $\alpha$=0.5, $L_{\text{max}}$=120, $\varphi$ =0.3 and $\lambda$=0.09. AdamW is used as the optimizer. All experiments are conducted on 8 NVIDIA H20 96GB GPUs. 

\begin{table}[h]
\scriptsize
\centering
\setlength{\tabcolsep}{0.33mm}
\begin{tabular}{lccc ccc}
\toprule
\multirow{2}{*}{\textbf{Method}} & \multicolumn{3}{c}{\textbf{AC--Low}} & \multicolumn{3}{c}{\textbf{AC--High}} \\
\cmidrule(lr){2-4} \cmidrule(lr){5-7}
 & Type & Ground. & SR & Type & Ground. & SR \\
\midrule
GPT-4o\cite{hurst2024gpt-4o} & 74.3 & 38.7 & 28.4 & 63.1 & 30.9 & 21.2 \\
OS-Genesis-7B\cite{sun2025genesis}& 90.7 & -- & 74.2 & 66.2 & -- & 44.5 \\
GUI-R1-3B\cite{guir1} & -- & -- & -- & 58.0 & 56.2 & 46.6 \\
GUI-R1-7B\cite{guir1} & -- & -- & -- & 71.6 & 65.6 & 51.7 \\
SeeClick\cite{cheng2024seeclick} & \underline{93.0} & 73.4 & 75.0 & 82.9 & 62.9 & 59.1 \\ 
InternVL-2-4B\cite{chen2024internvl} &90.9 & \underline{84.1} & 80.1 & \underline{84.1} & \underline{72.7} & \underline{66.7}\\
OS-Atlas-4B\cite{wu2024atlas} & 91.9 & 83.8 & \underline{80.6} & \textbf{84.7} & \textbf{73.8} & \textbf{67.5} \\
\midrule
\model & \textbf{97.2} & \textbf{86.7} & \textbf{92.1} & 77.1 & 72.6 & {65.5} \\
\bottomrule
\end{tabular}
\caption{Performance comparison on AndroidControl-Low (AC-Low) and AndroidControl-High (AC-High).}
\label{androidcontrol}
\end{table}
\subsection{Benchmarks}
\model\ is evaluated across web, mobile, and desktop environments. We use ScreenSpot~\cite{cheng2024seeclick}, ScreenSpot‑v2~\cite{li2025screenspot}, and ScreenSpot‑pro~\cite{screenspot-pro} to assess screen grounding, and AndroidControl~\cite{androidcontrol}, a static single‑step mobile benchmark, to assess agentic performance. To align real‑world usage, multi‑role orchestration is evaluated in the online web environment MiniWob++~\cite{miniwob} and the online mobile environment AndroidWorld~\cite{rawles2024androidworld}. In addition, OS‑World~\cite{xie2024osworld} is used to measure the effectiveness of \model\ as a policy executor in computer‑use online scenarios.
See App.~\ref{appix-metrics} for benchmark details and the corresponding metrics.

\subsection{GUI-Oriented Foundation Performance}
We evaluate the fundamental capabilities of \model\ in processing GUI-related tasks by screen grounding (Tab.\ref{screenspot-pro}) and step-wise agentic performance (Tab.\ref{androidcontrol}). As shown in Tab.\ref{screenspot-pro}, \model\ achieves overall leading performance, particularly against GUI-specialized methods with substantially larger parameter scales, such as UGround-72B, UI‑TARS‑7B, OS-Atlas-7B and UI‑S1‑7B. {Compared with the foundational model Qwen2.5‑VL-3B, our approach yields a 39.4\% overall gain.} 
Under comparable model sizes, \model\ consistently outperforms previous methods, and {surpasses the advanced methods on graphical UI element grounding}, especially in the challenging CAD scenarios. We observe that \model\ exhibits stable visual perception of small‑size UI elements, enabling reliable perception across screens with varying resolutions. The stable screen grounding provides a solid foundation for precise screen interaction in GUI automation. In Tab.\ref{androidcontrol}, compared with the methods tailored for single-task optimization on the training set, our \model\ still achieves competitive results. Particularly, the leading performance on AC-Low demonstrates that \model\ can achieve accurate action-tool alignment under explicit action instructions, verifying the effectiveness of the \framework\ framework. 

\begin{table}[h]
\scriptsize
\centering
\begin{tabular}{lc}
\toprule
\textbf{Method} & \textbf{Success Rate} \\
\midrule
Qwen2.5-VL-3B~\cite{bai2025qwen2.5} & 34.6 \\
OS-Atlas-7B~\cite{wu2024atlas} & 35.2 \\
AgentCPM-GUI-8B~\cite{zhang2025agentcpm} & 37.8 \\
Qwen2.5-VL-7B~\cite{bai2025qwen2.5} & 54.0 \\
UI-TARS-7B~\cite{qin2025ui} & 58.7 \\
UI-S1-7B~\cite{uis1} & 60.9 \\
Aguvis-72B~\cite{xu2024aguvis} & 66.0 \\
Gemini-2.5-pro~\cite{Gemini2025} & \underline{71.0} \\
\midrule
\rowcolor{mygray}
\multicolumn{2}{l}{\textit{End2End Reasoning}}\\
\model & 50.0 \\
\rowcolor{mygray}
\multicolumn{2}{l}{\textit{Multi-Agent System}}\\
\model & 60.9 {\scriptsize $(+21.8\%)$} \\
\rowcolor{mygray}
\multicolumn{2}{l}{\textit{Policy Executor Mode}}\\
\model\ (Gemini-2.5-pro as planner)  & \textbf{77.2} {\scriptsize $(+54.4\%)$} \\ 
\bottomrule
\end{tabular}
\caption{Performance on MiniWob++.}
\label{miniwob}
\end{table}
\subsection{Effectiveness of Multi-role Orchestration}
We evaluate the multi‑role orchestration of \model\ within the MiniWob++ and AndroidWorld.

\noindent\textbf{End2End Reasoning.} 
In Tab.~\ref{miniwob}, under end-to-end reasoning, \model\ outperforms Qwen2.5-VL-3B by $44.5\%$ and remains competitive with larger GUI agents, surpassing even larger GUI-specific baselines such as OS-Atlas-7B ($+42.0\%$) and AgentCPM-GUI-8B ($+32.3\%$). We attribute the gains to \textbf{\textit{(i)}} domain knowledge acquired during SFT, which enables \model\ to select appropriate policies conditioned on the current screen state; \textbf{\textit{(ii)}} multi-task exploration during RL, which helps \model\ adapt its policy and recover by exploring alternative pathways when execution deviates from the goal; and \textbf{\textit{(iii)}} PWCE training with enhanced UI-element semantics and ILG data, which jointly improve fine-grained screen perception and accurate target clicking. These results suggest the efficacy of our method in GUI episodic reasoning.
\begin{table}[h]
\scriptsize
\centering
\begin{tabular}{lc}
\toprule
\textbf{Method} & \textbf{Success Rate} \\
\midrule
Aguvis-72B~\cite{xu2024aguvis} & 26.1 \\
Claude Computer Use~\cite{claude} & 27.9\\
Gemini-2.5-pro~\cite{Gemini2025} & 31.0 \\
Qwen2.5-VL-32B~\cite{bai2025qwen2.5} & 31.5 \\
GPT-4o + Aria-UI~\cite{yang2025aria} & 44.8 \\
UI-TARS-72B~\cite{qin2025ui} & 46.6 \\
OpenAI CUA-o3~\cite{openai2025computeragent} & 52.5 \\
Agent-S2~\cite{agents2} & 54.3 \\
Qwen3-VL-235B-A22B-Instruct~\cite{qwen3vl} & 63.7 \\
UI-Venus-Navi-72B~\cite{uivenus} & 65.9 \\
JT-GUIAgentV2~\cite{chinmobile2025jtguiagent} & 67.2 \\
Mobile-Agent-V3~\cite{MobileAgentv3} & 73.3 \\ 
UI-Ins-7B~\cite{uiins} (GPT-5 as planner) & \underline{74.1} \\
\midrule
\rowcolor{mygray}
\multicolumn{2}{l}{\textit{Policy Executor Mode}}\\
\model\ (Gemini-2.5-pro as planner) & 60.3 \\ 
\model\ (GPT-5 as planner) & \textbf{77.6} \\ 
\bottomrule
\end{tabular}
\caption{Performance on AndroidWorld.}
\label{androidword}
\end{table}

\noindent\textbf{Multi-Agent System.}
In Tab.~\ref{miniwob}, orchestrating \model\ into a parameter-shared MAS with context engineering further improves performance by $21.8\%$. Compared with single-agent, MAS mitigates two common issues:
{(\textit{i}) Thought–Action Hallucination}: In end-to-end reasoning, a single agent must jointly perform goal analysis, screen perception, and tool invocation, \textit{which may lead to misalignment between reasoning and actions on OOD tasks.} MAS decomposes the overall goal into subtasks, reducing per-agent complexity and enabling low-hallucination collaboration.
{(\textit{ii}) Weak History Awareness}: As interactions grow longer, \textit{increasing context length exacerbates the “lost-in-the-middle” issue, often leading to action loops.} MAS manages context per role to keep inputs concise, improving the awareness of historical interaction clues and reducing repetitive failures.

\noindent\textbf{Policy Executor Mode.}
As shown in Tab.~\ref{miniwob}, when integrated with \model, our method achieves an $8.7\%$ gains over monolithic  Gemini‑2.5‑Pro\footnote{gemini-2.5-pro-preview-05-06}. Moreover, its attains leading performance: it outperforms Aguvis‑72B, a powerful end‑to‑end GUI agent, by $17.0\%$, and yields a substantial $54.4\%$ gains over the single-agent setting of \model. In Tab.~\ref{androidword}, {pairing Gemini-2.5-Pro as the planner with \model\ yields a 94.5\% relative improvement over Gemini-2.5-Pro in the end-to-end reasoning (31.0 to 60.3)}. This suggests that \model, when acting as a policy executor, can reliably execute low-level interactions on real devices. When equipped with a more advanced planner GPT-5\footnote{gpt-5-2025-08-07}, our framework achieves leading performance, surpassing previous SOTA methods by a considerable margin: {it yields a 5.9\% improvement over the MAS framework Mobile-Agent-V3, an 17.8\% improvement over the end‑to‑end native GUI agent UI‑Venus‑Navi‑72B, and exceeds the larger‑parameter executor UI‑Ins‑7B by 3.5 points.}
Taken together, these results indicate that  \textbf{(\textit{i})} \model\ can effectively integrate with advanced MLLMs, enabling collaborative GUI automation that more favorably trades off scalability against computational and resource costs; and \textbf{(\textit{ii})} compared with end‑to‑end monolithic models (e.g., UI‑TARS‑72B and UI‑Venus‑Navi‑72B), {\model, when serving as a policy executor in conjunction with continuously evolving MLLMs as planners, can attain a higher performance ceiling for GUI automation.}

\subsection{Policy Executor Capability Assessment}
Tab.~\ref{miniwob} and Tab.~\ref{androidword} demonstrate the potential of the policy executor mode for deploying lightweight MLLMs in GUI automation. To evaluate \model\ as a stable policy executor, we benchmark it on OS‑World (max. 50 steps) vs. advanced GUI-capable MLLMs using their official prompts, including generic models (Qwen2.5‑VL‑32B, Qwen3‑VL‑4B) and GUI‑specialized models (UI-TARS-1.5-7B, InfiGUI‑R1-3B). As shown in Tab.~\ref{executor_eval}, {\model\ outperforms Qwen3-VL-4B by 15.6\%, trails Qwen2.5-VL-32B by only 5.1 points with 10× fewer parameters, and surpasses the advanced method UI-TARS-1.5-7B by 36.5\% with substantially fewer parameters}. Notably, {InfiGUI-R1-3B with the same backbone is competitive on static environments yet drops by 28.2 points in online environments compared with \model}\footnote{Results highlight the constrained task scalability of training on end-to-end agentic episodes.}, underscoring \model's scalable execution as a robust policy executor in navigating realistic GUI workflows.

\begin{table}[]
  \centering
  \scriptsize
  \setlength{\tabcolsep}{1mm}
  \begin{tabular}{l l c}
    \toprule
    \textbf{Planner} & \textbf{Policy Executor} & \textbf{Success Rate} \\
    \midrule

    \multirow{8}{*}{Gemini-2.5-pro}
      & \multicolumn{2}{l}{\cellcolor{mygray}\textit{Generic Models}} \\
      & Qwen2.5-VL-32B~\cite{bai2025qwen2.5}   & 43.6   \\
      & Qwen3-VL-4B~\cite{qwen3vl}             & 33.3 \\

      & \multicolumn{2}{l}{\cellcolor{mygray}\textit{GUI-specific Models}} \\
      & UI-TARS-1.5-7B~\cite{ui-tars-15-seed} & 28.2 \\
      & InfiGUI-R1-3B~\cite{infigui-r1}    & 10.3   \\
    & \multicolumn{2}{l}{\cellcolor{mycolor}\textit{ours}}\\
      & \textbf{\model}                                        & 38.5 \\
    \bottomrule
  \end{tabular}
  \caption{Comparison of different executors in OSWorld. We adopt the official split of 39 tasks (see App.~\ref{appix-metrics}).}
  \label{executor_eval}
\end{table}

\begin{table}[]
  \centering
  \scriptsize
  \begin{tabular}{lcccc}
    \toprule
    \textbf{Method}           & \textbf{SP}              & \textbf{SP-v2}            & \textbf{SP-pro}             & \textbf{MW}          \\
    \midrule
    \rowcolor{mycolor}
    \textbf{\model}\         & \textbf{84.3}            & \textbf{86.4}             & \textbf{36.1}               & \textbf{50.0}               \\
    \shortstack{ --- \textit{w/o} ILG \\ $ $}    & \shortstack{82.6 \\ {\scriptsize(--2.1\%)}}   & \shortstack{83.2 \\ {\scriptsize(--3.8\%)}}    & \shortstack{26.8 \\ {\scriptsize(--34.7\%)}}     & \shortstack{48.7 \\ {\scriptsize(--2.7\%)}}   \\
    \midrule
    \shortstack{SFT stage \\ $ $}     & \shortstack{83.4 \\ {\scriptsize(--1.1\%)}}   & \shortstack{83.9 \\ {\scriptsize(--3.0\%)}}   & \shortstack{27.2 \\ {\scriptsize(--32.7\%)}}     & \shortstack{40.8 \\ {\scriptsize(--22.5\%)}} \\
    \shortstack{ --- \textit{w/o} PWCE \\ $ $}           & \shortstack{82.9 \\ {\scriptsize(--1.7\%)}}   & \shortstack{83.5 \\ {\scriptsize(--3.5\%)}}  & \shortstack{26.1 \\ {\scriptsize(--38.3\%)}}    & \shortstack{39.4 \\ {\scriptsize(--26.9\%)}}     \\
    \midrule
    \rowcolor{mygray}
    \shortstack{Qwen2.5-VL-3B \\ $ $}   & \shortstack{78.3 \\ {\scriptsize(--7.7\%)}}   & \shortstack{81.3 \\ {\scriptsize(--6.3\%)}}    & \shortstack{23.9 \\ {\scriptsize (--51.0\%)}}    & \shortstack{34.6 \\ {\scriptsize (--44.5\%)}}     \\  
    \bottomrule
  \end{tabular}
  \caption{Ablation results on ScreenSpot (SP), ScreenSpot-v2 (SP-v2),
  ScreenSpot-pro (SP-pro), and MiniWob++ (MW). Numbers in parentheses indicate the relative performance drop (\%) vs. \model.}
  \label{ablation}
\end{table}

\subsection{Ablation Study}

In Tab.~\ref{ablation}, we evaluate key designs of \framework.

\noindent \textbf{Contributions of the two‑round recipe.}
Starting from Qwen2.5‑VL‑3B, the full two‑round training recipe yields \model, improving SP/SP‑v2/ SP‑pro/MW by $7.7\%$/$6.3\%$/$51.0\%$/$44.5\%$, respectively. The SFT stage contributes an average $10.4\%$ gains, and the subsequent RL stage brings an additional $14.8\%$ gains. Indicating that our SFT recipe effectively equips the GUI agent with domain knowledge, while the RL recipe further teaches it to explore optimal policies for GUI‑related tasks.

\noindent \textbf{Effectiveness of PWCE.}
Within the SFT stage, replacing PWCE with vanilla cross‑entropy loss $\mathcal L_{CE}$ leads to an average $2.2\%$ degradation in GUI grounding and agentic performance, including $4.2\%$ on SP-pro (with complex, high‑resolution layouts) and $3.6\%$ on MW. This indicates that PWCE provides more effective supervision, encouraging the GUI agent to learn fine‑grained visual clues.

\noindent \textbf{Effectiveness of ILG data.}
Since ILG data is used in the RL stage to equip the GUI agent with stable grounding ability under intricate screen layouts, {removing it results in a 34.7\% degradation on the SP-pro} and a $2.7\%$ drop in agentic performance, which relies on accurate low-level GUI interaction (e.g., \texttt{click}, \texttt{long\_press}, and \texttt{swipe} operations).

%% file: conclusion.tex
\section{Conclusion}
In this work, we focus on enhancing lightweight MLLMs to participate in realistic GUI workflows and propose \framework\ to equip them with robust task scalability for expanding their capability boundary to solve increasingly complex in-the-wild scenarios. Via \framework, we develop \model, a task-scalable lightweight GUI agent. When functioning as a policy executor, \model\ enables precise low-level GUI execution and, paired with advanced planners, it continually benefits from advances in planning, offering a higher performance ceiling. Experiments in both static and online configurations validate the effectiveness of our designs.

%% file: appendix.tex
\section{Appendix}
\subsection{Data Curation}\label{appix-datacuration}

We curate agentic data from GUI-oriented datasets across mobile and desktop platforms, sampling 160k mobile instances from Aguvis~\cite{xu2024aguvis} (AMEX~\cite{amex}, GUI-Odyssey~\cite{odyssey}, AndroidControl~\cite{androidcontrol}, AITW~\cite{aitw}) and 140k PC/Web instances from AgentNet~\cite{opencua}. From AgentNet’s metadata, we directly extract action--tool alignment, CoT reasoning, and screen summarization, and standardize them into unified training pairs.

For Goal Planning (GP) synthesis, we select 18k episodes from the curated metadata and employ Gemini-2.5-pro\footnote{gemini-2.5-pro-preview-05-06} to generate tailored planning descriptions for each goal. For Screen Understanding (SU) synthesis, 20k mobile screenshots are processed with Gemini-2.5-pro to produce detailed textual descriptions. For Logic-Consistent CoT (LCC) synthesis, we sample 40k mobile items and use Gemini-2.5-pro to generate step-wise reasoning aligned with each action\footnote{The original System-1 direct output style is transformed into a System-2 CoT reasoning format.}.  
For Action–Tool Alignment (ATA) synthesis, 60k mobile samples are processed with Qwen2.5-VL-72B-Instruct to produce functional descriptions of actions in the current screen state. For Screen Grounding (SG) data, we sample 30k items from OS-ATLAS~\cite{wu2024atlas} and employ Qwen2.5-VL-72B-Instruct to rewrite their original element instructions into semantically rich descriptions. Additionally, 6k samples serve as seeds for the rule-based data augmentation pipeline, generating 20k high-resolution grounding (ILG) instances with intricate layouts.  

In total, we obtain approximately 500k hybrid training samples, of which 400k are allocated for round-1 SFT and 100k (including the 20k ILG samples) are reserved for round-2 RL.

\subsection{Benchmark Details and Metrics}\label{appix-metrics}

\noindent\textbf{ScreenSpot}~\cite{cheng2024seeclick}:
ScreenSpot is a GUI grounding benchmark with 1,200+ instructions from iOS, Android, macOS, Windows, and Web screenshots, with targets annotated as either \textit{Text} or \textit{Icon/Widget}. It evaluates a core capability required by real-world GUI agents: accurately translating user language into the correct on-screen interaction target, which directly impacts usability and safety. 
\emph{\textbf{Metric:}} Accuracy, computed as whether the predicted coordinate point falls within the ground-truth bounding box.

\noindent\textbf{ScreenSpot-v2}~\cite{li2025screenspot}:
ScreenSpot-v2 is an upgraded GUI grounding benchmark that reduces annotation ambiguity and provides clearer instruction-to-target mappings across mobile/desktop/web screenshots.
\emph{\textbf{Metric:}} Accuracy.

\noindent\textbf{ScreenSpot-pro}~\cite{screenspot-pro}:
ScreenSpot-pro targets professional, high-resolution software interfaces and evaluates grounding on complex applications. It contains 1,581 instructions over 23 applications in 5 categories (development tools, creative apps, CAD/engineering, scientific/analytical, and office software) across Windows/macOS/Linux.
\emph{\textbf{Metric:}} Accuracy.

\noindent\textbf{AndroidControl}~\cite{androidcontrol}:
AndroidControl is a large-scale Android computer-control dataset with 15,283 human demonstrations spanning 833 apps, where each instance provides both high-level (episode-wise, AC-High) and low-level (step-wise, AC-Low) goals. \emph{\textbf{Metrics:}} Step-wise results. Type accuracy (correct action type), Grounding accuracy (for coordinate-based actions such as click/longPress), and step-wise success rate (SR), where both action type and action value must match the ground truth at each step.

\noindent\textbf{MiniWob++}~\cite{miniwob}:
MiniWob++ is an online benchmark that provides 100+ web interaction environments with Gymnasium-style interfaces, enabling controlled, programmatic evaluation of browser-based web automation via Selenium. Its real-world value is that many everyday tasks—filling forms, clicking buttons, navigating pages—are web-based. We use the test environment provided by \citealp{rawles2024androidworld}, comprising a total of 92 online tasks.
\emph{\textbf{Metric:}} Episode-wise success rate (SR), i.e., whether the agent completes the user goal by the end of the episode, judged by environment/state checks.
\begin{table*}[]
\centering
\footnotesize
\setlength{\tabcolsep}{6pt}
\begin{tabular}{ll}
\toprule
\rowcolor{mygray}
\multicolumn{2}{@{}c@{}}{\textbf{PC/Web Environment}} \\
\textbf{Atomic Action} & \textbf{Description} \\
\midrule
\texttt{pyautogui.click(x=x1, y=y1)} &
Click/Tap the UI element at screen coordinate $(x_1, y_1)$. \\
\texttt{pyautogui.doubleClick(x=x1, y=y1)} &
Double-click at $(x_1, y_1)$. \\
\texttt{pyautogui.rightClick(x=x1, y=y1)} &
Right-click at $(x_1, y_1)$. \\
\texttt{pyautogui.moveTo(x=x1, y=y1)} &
Move the cursor to $(x_1, y_1)$. \\
\texttt{pyautogui.dragTo(x=x1, y=y1)} &
Drag the cursor to $(x_1, y_1)$. \\
\texttt{pyautogui.press(keys=['key'])} &
Press a keyboard key. \\
\texttt{pyautogui.hotkey(keys=['key1','key2',...])} &
Press a keyboard shortcut combination. \\
\texttt{pyautogui.write(message='text')} &
Type a text string. \\
\texttt{pyautogui.scroll(direction)} &
Scroll the mouse wheel in a specified direction (\texttt{up/down/left/right}). \\
\texttt{pyautogui.wait()} &
Wait for loading. \\
\texttt{pyautogui.terminate(status='success')} &
Terminate the episode when the goal is achieved. \\
\midrule
\rowcolor{mygray}
\multicolumn{2}{@{}c@{}}{\textbf{Mobile Environment}} \\
\textbf{Atomic Action} & \textbf{Description} \\
\midrule
\texttt{pyautogui.click(x=x1, y=y1)} &
Click/Tap the UI element at screen coordinate $(x_1, y_1)$. \\
\texttt{mobile.long\_press(x=x1, y=y1)} &
Long-press at $(x_1, y_1)$. \\
\texttt{pyautogui.press(keys=['enter'])} &
Press the \texttt{Enter} key. \\
\texttt{mobile.swipe(begin=[x1,y1], end=[x2,y2])} &
Swipe from $(x_1, y_1)$ to $(x_2, y_2)$. \\
\texttt{pyautogui.wait()} &
Wait for loading. \\
\texttt{pyautogui.terminate(status='success')} &
Terminate the episode when the goal is achieved. \\
\texttt{pyautogui.write(message='text')} &
Type a text string. \\
\texttt{mobile.open\_app(name='APP name')} &
Open the app with the specified name. \\
\texttt{pyautogui.answer(message='text')} &
Provide a text answer to the user. \\
\texttt{mobile.home()} &
Go to the home screen. \\
\texttt{mobile.back()} &
Press the back button. \\
\bottomrule
\end{tabular}
\caption{Action space for LAMO-3B.}
\label{tab:action-space}
\end{table*}

\noindent\textbf{AndroidWorld}~\cite{rawles2024androidworld}:
AndroidWorld is a real-world-aligned simulation environment developed with Android Studio\footnote{https://developer.android.com/studio} and is widely used as an online benchmark for autonomous agents in Android settings, featuring 116 tasks spanning 20 real-world apps.
\emph{\textbf{Metric:}} Episode-wise SR.

\noindent\textbf{OSWorld}~\cite{xie2024osworld}:
OSWorld is a scalable real-computer environment for multimodal agents across operating systems (Ubuntu, Windows, and macOS), supporting task setup and execution-based evaluation. It provides a benchmark of 369 tasks spanning real web and desktop applications, OS-level file I/O, and cross-application workflows. Each task includes an initial-state configuration and an evaluation script to ensure reproducibility. Given the huge token cost of OSWorld’s large-scale task set, and since our goal is to assess whether \model\ can accurately execute low-level GUI interaction in computer-user scenarios, we adopt the official split of 39 tasks (category-wise sampled from the original 369 tasks), covering 10 computer-use domains, including office, daily, and professional settings (Tab.~\ref{tab:domain_task_count}). 
\emph{\textbf{Metric:}} Episode-wise SR.

\subsection{\model\ Action Space}\label{appix-instruction}
Table~\ref{tab:action-space} presents the rich action space supported by the \model\, encompassing commonly used atomic actions on both Mobile and PC platforms. These actions are parsed, in the form of tool calls, into interaction codes supported by the corresponding environment (\texttt{Android Debug Bridge}\footnote{https://developer.android.com/tools/adb} in the mobile environment and \texttt{pyautogui}\footnote{https://pyautogui.readthedocs.io/en/latest/} on PC), thereby enabling direct manipulation of the devices.
\begin{table}[h]
\footnotesize
\centering
\begin{tabular}{c c}
\hline
\textbf{Domain} & \textbf{Tasks} \\
\hline
chrome & 4 \\
gimp & 2 \\
libreoffice\_calc & 3 \\
libreoffice\_impress & 2 \\
libreoffice\_writer & 2 \\
multi\_apps & 17 \\
os & 2 \\
vs\_code & 3 \\
thunderbird & 2 \\
vlc & 2 \\
\hline
\textbf{Total} & 39 \\
\hline
\end{tabular}
\caption{Statistics of the OSWorld official split small-scale test tasks.}
\label{tab:domain_task_count}
\end{table}

\begin{algorithm*}[ht]
\footnotesize
\caption{ILG data augmentation strategy}
\label{alg-ILG}
\textbf{Input:} 
meta sample $(o_i, \mathcal P^i_{\text{point}}, \mathcal C_{\text{orig}})\in \mathcal D^+$,
background view $\mathcal O_{\text{back}}\in \mathcal D^-$,
distractor screen list $\mathcal Y_{\text{distractor}}\in \mathcal D^+$ and $(o_i, \mathcal P^i_{\text{point}}, \mathcal C_{\text{orig}})\in\mathcal Y_{\text{distractor}}$
\\
\textbf{Output}: ILG sample $(\ddot{O}, \ddot{\mathcal P}^i_{\text{point}}, \mathcal C_{\text{orig}})$
\begin{algorithmic}[1]
\STATE $\dot{\mathcal O} \gets {background\_enhance}(\mathcal O_{\text{back}}, o_i)$ {\textit{// rotation/stitching/scaling}}
\FOR{$\mathcal O_{\text{distractor}} \ in \ \mathcal Y_{\text{distractor}}$}
\STATE $\ddot{\mathcal O} \gets {interference\_insert}(\dot{\mathcal O}, \mathcal O_{\text{distractor}})$
\ENDFOR

\STATE $\ddot{\mathcal P}^i_{\text{point}} \gets {coordinate\_scale}(\mathcal P^i_{\text{point}},\,\ddot{O})$

\RETURN ($\ddot{O}, \ddot{\mathcal P}^i_{\text{point}}, \mathcal C_{\text{orig}}$)
\end{algorithmic}
\end{algorithm*}

\subsection{ILG Data Augmentation.}\label{appix-ILG}
\begin{figure*}
    \centering
    \includegraphics[width=1.\linewidth]{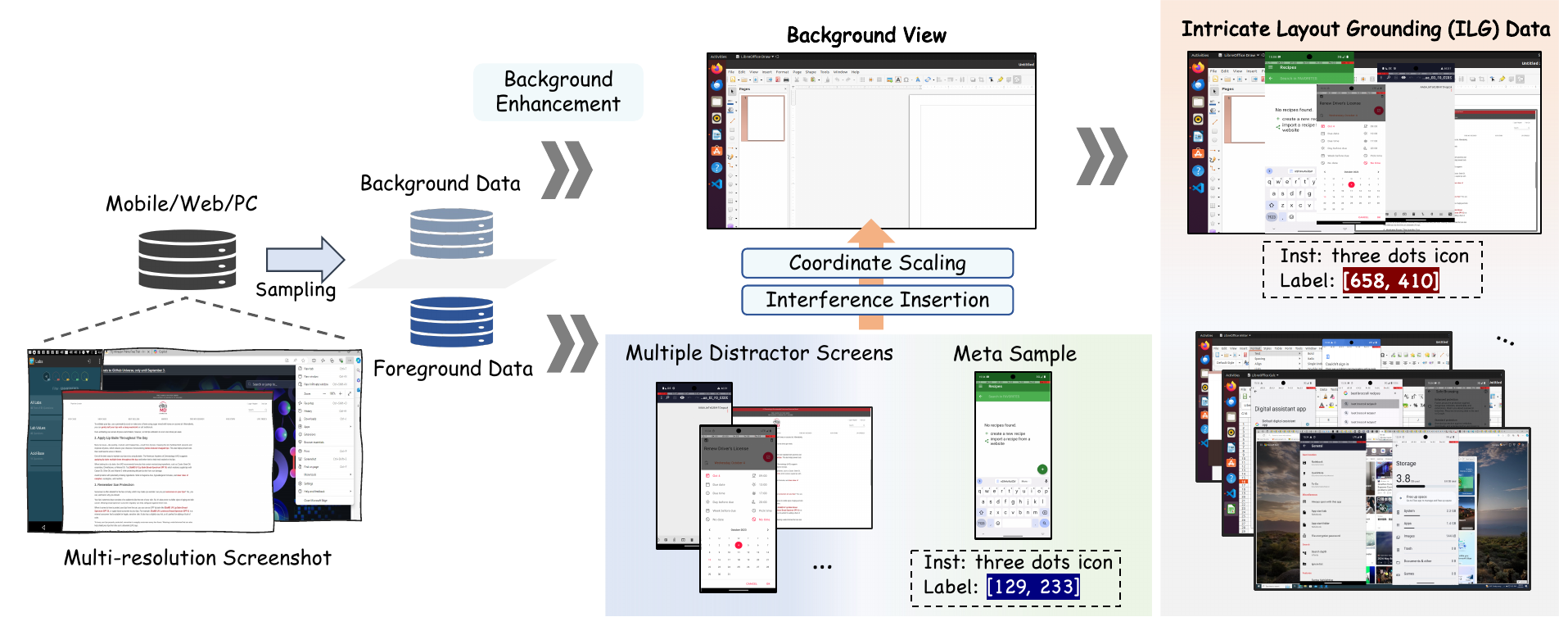}
    \caption{A toy example of the ILG data augmentation workflow.}
    \label{fig:ILG_workflow}
\end{figure*}

Algorithm~\ref{alg-ILG} provides detailed specifications of our ILG data augmentation (see Section~\ref{sec-role-oriented-data}), and Figure~\ref{fig:ILG_workflow} presents a toy example visualizing the ILG data augmentation workflow. With this approach, we can automatically synthesize large-scale, high-resolution screen-grounding data with complex layout variations, enhancing GUI agents’ fine-grained visual perception and their capability to operate on high-resolution screens.

\subsection{Performance on ScreenSpot and ScreenSpot-v2}\label{appix-grounding}
We report detailed results for \model\ on the ScreenSpot~\cite{cheng2024seeclick} and ScreenSpot-v2~\cite{li2025screenspot} screen-grounding benchmarks (Tabs.~\ref{tab:screenspot} and~\ref{tab:screenspot_v2}). \model\ consistently outperforms parameter-matched GUI-specialized baselines, with especially strong performance in mobile settings for grounding both textual and graphical UI elements. These results demonstrate \model's robust screen-perception capability, supporting accurate pixel-level GUI interactions whether it performs end-to-end reasoning (Fig.~\ref{fig:miniwob_single}) or serves as a policy executor (Figs.~\ref{fig:miniwob_multi}, ~\ref{fig:GPT5-AndroidWorld} and~\ref{fig:osworld-episode}).

\begin{table}[t]
\centering
\scriptsize
    \setlength{\tabcolsep}{1mm}
    \resizebox{\linewidth}{!}{
    \begin{tabular}{@{}lccccccc@{}}
        \toprule
         \multirow{2}{*}{\textbf{Method}} & \multicolumn{2}{c}{\textbf{Mobile}} & \multicolumn{2}{c}{\textbf{Desktop}} & \multicolumn{2}{c}{\textbf{Web}} &  \multirow{2}{*}{Avg.} \\
        \cline{2-3} \cline{4-5} \cline{6-7}
        & Text & Icon & Text & Icon & Text & Icon & \\
        \midrule
        GPT-4o~\cite{hurst2024gpt-4o} & 30.5 & 23.2 & 20.6 & 19.4 & 11.1 & 7.8 & 18.8 \\
        CogAgent~\cite{cogagent}&67.0 &24.0 &74.2 &20.0 &70.4 &28.6 &47.4\\
        SeeClick~\cite{cheng2024seeclick}&	78.4	&50.7	&70.1	&29.3	&55.2	&32.5&	55.1 \\
        R-VLM~\cite{r-vlm} & 85.0 &61.1 &81.4 &52.8 &66.5& 51.4 &66.3\\
        MP-GUI~\cite{wang2025mp} & 86.8 & 65.9 & 70.8 & 56.4 & 58.3 & 46.6 & 64.1 \\
        UGround-7B~\cite{gou2024navigating} & 82.8 & 60.3 & 82.5 & 63.6 & 80.4 & 70.4 & 73.3 \\
        ShowUI~\cite{lin2025showui} & 92.3 & \underline{75.5} & 76.3 & 61.1 & 81.7 & 63.6 & 75.1 \\
        UI-R1-3B~\cite{lu2025ui} & -- & -- & 90.2 & 59.3 & 85.2 & 73.3 & -- \\
        GUI-R1-3B~\cite{guir1} & -- & -- & 93.8 & 64.8 & \underline{89.6} & 72.1 & -- \\
        UI-TARS-2B~\cite{qin2025ui} & 93.0 & 75.5 & \underline{94.3} & 68.6 & 84.3 & \textbf{74.8} & 82.3 \\
        OS-Atlas-7B~\cite{wu2024atlas}  & 93.0 & 72.9 & 91.8 & 62.9 & \textbf{90.9} & \underline{74.3} & 82.5 \\
        HAR-GUI-3B~\cite{hargui} & \underline{94.5} & \underline{81.0} & 93.8 & \underline{70.8} & 85.6 & 73.8 & \underline{83.3} \\
        \midrule
        \model\ & \textbf{97.1} & \textbf{81.7} & \textbf{95.4} & \textbf{72.1} & 86.2 & 73.1 & \textbf{84.3} \\
        \bottomrule
    \end{tabular}
    }
    \caption{Performance comparison on ScreenSpot.}
    \label{tab:screenspot}
\end{table}
\begin{table}[h]\scriptsize
\centering
    \setlength{\tabcolsep}{0.3mm}
    \begin{tabular}{lccccccc}
        \toprule
        \multirow{2}{*}{\textbf{Method}} & \multicolumn{2}{c}{\textbf{Mobile}} & \multicolumn{2}{c}{\textbf{Desktop}} & \multicolumn{2}{c}{\textbf{Web}} & \multirow{2}{*}{Avg.} \\
        \cline{2-3} \cline{4-5} \cline{6-7}
        & Text & Icon & Text & Icon & Text & Icon & \\ 
        \midrule
        SeeClick~\cite{cheng2024seeclick} & 78.4 & 50.7 & 70.1 & 29.3 & 55.2 & 32.5 & 55.1 \\
        GPT-4o + SeeClick~\cite{cheng2024seeclick} & 85.2 & 58.8 & 79.9 & 37.1 & 72.7 & 30.1 & 63.6 \\
        OS-Atlas-4B~\cite{wu2024atlas} & 87.2 & 59.7 & 72.7 & 46.4 & 85.9 & 63.1 & 71.9 \\
        GPT-4o + OS-Atlas-4B~\cite{wu2024atlas} & 95.5 & 75.8 & 79.4 & 49.3 & 90.2 & 66.5&  79.1 \\
        InternVL3-8B~\cite{zhu2025internvl3}& -- &-- &-- & --& --&-- &  81.4\\ 
        Qwen2.5-VL-3B~\cite{bai2025qwen2.5} & 95.0	&80.1	&90.2&	64.3&	88.0&	70.4	&81.3\\
        OS-Atlas-7B~\cite{wu2024atlas} & 95.2 & 75.8 & 90.7 & 63.6 & \textbf{90.6} & \underline{77.3} & 84.1 \\
        UI-TARS-2B~\cite{qin2025ui} & 95.2 & 79.1 & 90.7 & 68.6 & \textbf{90.6} & \underline{77.3} & 84.7 \\
        UI-R1-3B~\cite{lu2025ui} &96.2	&\textbf{84.3}	&92.3	&63.6	&\underline{89.2} &	75.4 & 85.4\\
        HAR-GUI-3B~\cite{hargui} & \underline{96.5} & 81.0 & \underline{95.4} & \underline{76.5} & 88.8 & \textbf{78.8} & \underline{86.2}\\
        \midrule
        \model\ & \textbf{98.6} & \underline{83.7} & \textbf{96.1} & \textbf{76.8} & 89.1 & 74.2 & \textbf{86.4} \\
        \bottomrule
    \end{tabular}
    \caption{Performance comparison on ScreenSpot-v2.}
    \label{tab:screenspot_v2}
\end{table}

\subsection{Prompt Templates}\label{appix-instruction}
We list the instructions used for data synthesis (Section~\ref{sec-role-oriented-data}) and multi-role orchestration (Section~\ref{sec-mas}).

\begin{table*}[]
\centering
\begin{minipage}{\textwidth}
\begin{tcolorbox}[colback=gray!8, title = {The Prompt for SU Data Synthesis ($\mathcal I_{\text{S2W}}$)}]
{\ttfamily \footnotesize You are an expert AI assistant specialized in User Interface (UI) analysis and description. Your primary task is to generate a detailed, structured, and objective summary of on-screen content from a provided description of a mobile, desktop, or web screenshot.\\
These summaries will be used as high-quality training data. Therefore, accuracy, detail, and consistency are paramount. You must act as if you are "seeing" the screen described and meticulously document its contents.\\

Please keep the following output format:\\
<screen2word>Screen description content</screen2word>\\

}
\end{tcolorbox}
\end{minipage}
\end{table*}

\begin{table*}[]
\centering
\begin{minipage}{\textwidth}
\begin{tcolorbox}[colback=gray!8, title = {The Prompt for SG Data Synthesis ($\mathcal I_{\text{G}}$)}]
{\ttfamily \footnotesize Here is a low-level description of a UI element and its coordinates.\\

Element low-level description: \{DESCRIPTION\}\\

Coordinate: <point>\{COORDINATES\}</point>\\

Please generate a more specific, semantically richer high--level description--for example, describe an element from the perspective of its spatial context—to help the model accurately understand and locate it.\\

}
\end{tcolorbox}
\end{minipage}
\end{table*}

\begin{table*}[]
\centering
\begin{minipage}{\textwidth}
\begin{tcolorbox}[colback=gray!8, title = {The Prompt for ATA Data Synthesis ($\mathcal I_{\text{Tool}}$)}]
{\ttfamily \small You are an expert in GUI action description. You look at a screenshot and translate a low-level atomic action (e.g., click coordinates, typing text) into a precise, high-level, human-readable instruction.\\

You will be given:\\
1) **Screenshot**: the current screenshot image.\\
2) **Action Space**: a list of available action tools and their descriptions.\\
3) **Atomic Action**: the low-level action to be executed .\\

Here are some instructions for you:\\
**1) Ground the target**\\
- Identify the most likely UI element/region corresponding to the action.\\
- Use the screenshot to name the element (button / icon / tab / input box / list item / link).\\
- Add distinguishing clues: visible text, icon meaning, color, shape, and **spatial context (e.g., "to the right of the search bar", "top-right corner", "below the title").\\
- If multiple candidates exist, choose the best-supported one and disambiguate with context.\\

**2) Describe intent, not coordinates**\\
- Do \textbf{not} mention raw coordinates.\\
- Convert the action into a natural-language description of what the agent is doing on the UI.\\

**3) Be specific and semantically rich**\\
- Prefer "Tap the blue Search button to the right of the query field" over "Tap Search".\\
- For \texttt{write(...)}, specify \emph{where} the text goes (which input field) and what it will achieve.\\

Here is a sample for your reference:\\
**Atomic Action**:  pyautogui.click(x=268, y=439)\\
**Description**: "Click the magnifying-glass icon at the top-right of the screen to start searching."\\

=======\\
**Action Space**: \{ACTION\_SPACE\}\\
**Atomic Action**: \{ATOMIC\_ACTION\}\\

Please keep the following JSON output format:\\
\{ "tool\_call": Atomic Action, "semantic\_description": "one- or two-sentence grounded description of this action"\} \\

}
\end{tcolorbox}
\end{minipage}
\end{table*}

\begin{table*}[]
\centering
\begin{minipage}{\textwidth}
\begin{tcolorbox}[colback=gray!8, title = {The Prompt for LCC Data Synthesis ($\mathcal I_{\text{CoT}}$)}]
{\ttfamily\small
Your task is to generate only the **Thought** (a long chain of thought), which explains why the current step should execute the given **ATOMIC ACTION**, and outputs the reasoning process from task history to the current state.\\

You will be given:\\
1) **Previous Actions**: A list of actions that have been taken so far.
2) **Former thought**: A description of the thought process of the previous action. 
3) **Goal**: A description of the task to be accomplished.
4) **ATOMIC ACTION**: The predicted next action, including operation type and parameters, in pyautogui format.  
5) **Full Screenshot**: A screenshot showing the current state.\\

**Previous Actions**:
\{PREVIOUS\_ACTIONS\}  \\
**Former Thought**:
\{THE\_THOUGHT\_PROCESS\_OF\_THE\_PREVIOUS\_STEP\}  \\
**Goal**:
\{GOAL\}  \\
**ATOMIC ACTION**:\{CURRENT\_ATOMIC\_ACTION\}  \\

Please consider the following constraints when you are thinking.
- **State changes:**\\
  - Based on the current screenshot, naturally continue and adjust from the most recent action. This part should connect smoothly with the later reasoning, like self-talk.  \\
- **Memory:**\\
  - Add necessary information according to the history, 'former thought', and current screenshot.  \\
- **Step by Step assess the progress towards completing the task:**\\
  - Analyze which parts of the task have already been completed and how they contribute to the overall 'goal'.  \\
  - Make a plan or adjust the former plan on how to complete the task based on the history and current screenshot.  \\
- **Propose the logical next action:**\\
  - List the possible next actions that could advance the task from the current state.  \\
  - Evaluate these possible actions based on the current state and 'Previous Actions'.  \\
  - The logical next action must match the current full screenshot and be consistent with the type of action in the 'ATOMIC ACTION' and explain why.  \\
  - Anticipate how the system state will change after executing this action.  \\
- Only describe the logical next action. Never mention **ATOMIC ACTION**.\\
- Do not say things like “the predicted action shows…”. Express it as if I am reasoning naturally.\\
- In the thought process, never mention the mouse position in the image.\\
- Write in **first person**, as if I am speaking to myself.  \\

**Important Notes**:\\
1. Your principle: Your task is to guide that model to generate the **Thought** based only on the **Goal**, **Previous Actions**, **ATOMIC ACTION** and the Screenshot.\\
2. For mouse-related actions:\\
   (i)Ignore the mouse position in the screenshot. (ii)Do not infer the target from the mouse position. (iii) Mouse position is irrelevant and provides no valid clue.\\
3. For text editing or input-related actions:\\
   (i)Observe the cursor position to understand where the user is preparing to type or edit text. (ii)Merge repetitive actions (such as multiple spaces, backspaces, deletes, or enters) into one description, and specify the exact number. (iii) Infer the user’s true intent and predict what the final text will look like after the action is completed. (iv) The instruction should reflect the FINAL STATE of the text, not intermediate steps.\\
4. **Extremely important**: The output must contain only logical, executable instructions derived from the 'goal', task context, and action history. Do not mention any predicted action or mouse position.\\
5. **Extremely important**: The output Thought must be written as a single continuous paragraph of reasoning, like natural self-talk, not divided into bullet points or numbered sections.\\
You must strictly follow this structure in your answer.\\

======\\
OUTPUT\\
**Thought**: Output your rigorous and comprehensive thinking process.\\

}
\end{tcolorbox}
\end{minipage}
\end{table*}

\begin{table*}[]
\centering
\begin{minipage}{\textwidth}
\begin{tcolorbox}[colback=gray!8, title={The Prompt for GP Data Synthesis ($\mathcal{I}_{\text{Plan}}$)}]
{\ttfamily\small
You are an expert in GUI task decomposition. You analyze user goals and visual evidence (screenshots) to create high-level strategic plans for GUI agents.

**Goal:** \{GOAL\}\\
**Visual Trajectory:** A sequence of screenshots is provided, showing the successful completion of the user's goal on a device.\\

Here are some instructions for you:\\
**Generate a High-Level Plan:** Create a multi-step plan for finishing the user's goal. Each step should describe the intent of a major phase in the task (e.g., "Search for the item," "Configure product options," "Enter shipping details"). Do not describe low-level actions like "Click the button at coordinates (X, Y)" or "Type the letter `A`."\\
**Ground the Plan:** Ensure every step in your plan corresponds to a logical segment (single or multiple screenshots) of the provided screenshot sequence.\\

Here is a sample for your reference:\\
\{\!"Goal": "Go to McDonald's and order a Big Mac and have it delivered to the address: xxx",
"Planning": "First, access the McDonald's app/mini program, find the Big Mac within the app, select the product attributes based on your needs, and finally, fill in your address in the address bar to complete the order.",
"Tips": "During this process, you should remember the following: (1) Pay attention to your historical operation history to avoid repeatedly clicking the same area; (2) Wait for the page to fully load before proceeding; (3) ..."\!\}\\

Please keep the following JSON output format:\\
\{\!"Goal": "**Goal:**",
"Planning": "Use a paragraph to break down the user's goals into a logical, high-level plan.",
"Tips": "Summarize the tips you think the user needs to pay attention to in completing this task."\!\} \\

}
\end{tcolorbox}
\end{minipage}
\end{table*}

\begin{table*}[]
\centering
\begin{minipage}{\textwidth}
\begin{tcolorbox}[colback=gray!8, title = {The Instruction for End2End Reasoning ($\mathcal I_{\text{e2e}}$)}]
{\tt \small You are given a goal and a screenshot. You need to perform a series of actions to complete the goal.\\
Here are the tools you can use: \{ACTION\_SPACE\}\\

Now, please generate the next action according to the goal:\\
\{"goal":\{GOAL\}, "Interaction History": \{HISTORY\}\} \\

Please keep the following format:\\
<think>analyze the current screen status step-by-step to plan the current interaction</think>\\
<action>a brief description of the current action</action>\\
<tool\_call>determine the execution tool/tools based on the current action</tool\_call>\\

}
\end{tcolorbox}
\end{minipage}
\end{table*}
\begin{table*}[]
\centering
\begin{minipage}{\textwidth}
\begin{tcolorbox}[colback=gray!8, title={Prompt for Screen Grounding ($\mathcal I^*_{G}$)}]
{\ttfamily\footnotesize
Locate the element on the screen with the function or description: \{ELEMENT\_DESCRIPTION\}.\\
Keep the following output format: \{"point\_2d": [x, y], "label": "re-describe the element to help you grounding"\}.\\

}
\end{tcolorbox}
\end{minipage}
\end{table*}
\begin{table*}[]
\centering
\begin{minipage}{\textwidth}
\begin{tcolorbox}[colback=gray!8, title = {The Instruction for MAS-Obesrver ($\mathcal{M}_\Theta, \, \tilde{\mathcal{I}}_{\mathrm{obs}}\longrightarrow \mathcal{M}_\Theta^{\mathrm{Obesrver}}$)}]
{\tt \small You are an expert AI assistant specialized in User Interface (UI) analysis and description. Your primary task is to generate a detailed, structured, and objective summary of on-screen content from a provided description of a mobile, desktop, or web screenshot.\\
Therefore, accuracy, detail, and consistency are paramount. You must act as if you are "seeing" the screen described and meticulously document its contents.\\
Please keep the following output format:\\
<screen2word>Screen description content</screen2word>\\

}
\end{tcolorbox}
\end{minipage}
\end{table*}

\begin{table*}[]
\centering
\begin{minipage}{\textwidth}
\begin{tcolorbox}[colback=gray!8, title = {The Instruction for MAS-Planner ($\mathcal{M}_\Theta, \, \tilde{\mathcal{I}}_{\mathrm{plan}}\longrightarrow \mathcal{M}_\Theta^{\mathrm{Planner}}$)}]

{\tt \small You are an expert in GUI task decomposition. Your task is to analyze user's goal and initial screenshot to create high-level strategic plans for GUI agents.\\
- Here is user's goal: \{GOAL\}\\

Please keep the following output format:\\
\{"Planning": Use a paragraph to break down the user's goals into a logical, high-level plan., "Tips": Summarize the tips you think the user need to pay attention to in completing this task.\}\\

}
\end{tcolorbox}
\end{minipage}
\end{table*}

\begin{table*}[]
\centering
\begin{minipage}{\textwidth}
\begin{tcolorbox}[colback=gray!8, title = {The Instruction for MAS-Allocator ($\mathcal{M}_\Theta, \, \tilde{\mathcal{I}}_{\mathrm{act}}\longrightarrow \mathcal{M}_\Theta^{\mathrm{Allocator}}$)}]
{\tt \small Please determine the action that should be taken now, based on the current screen state, **Interaction History**, **Observation**, and **Task Planning**.\\

**Observation**\\
<observation>\{SCREEN\_DESCRIPTION\}</observation>\\
**Task Planning**\\
<plan>\{PLAN\}</plan>\\
**Interaction History**\\
<history>\{HISTORY\}</history>\\

-- Here are helpful tips: \{TIPS\}\\
Please keep the following output format:\\
<action>decide what action should to be taken currently</action>"\\

}
\end{tcolorbox}
\end{minipage}
\end{table*}

\begin{table*}[]
\centering
\begin{minipage}{\textwidth}
\begin{tcolorbox}[colback=gray!8, title = {The Instruction for MAS-Executor ($\mathcal{M}_\Theta, \, \tilde{\mathcal{I}}_{\mathrm{exec}}\longrightarrow \mathcal{M}_\Theta^{\mathrm{Executor}}$)}]
{\tt \small You are given an instruction and a screenshot. You need to perform one or more actions to align the instruction.
Here are the tools you can use: \{ACTION\_SPACE\}\\

Instruction: \{ACTION\}\\

Please keep the following output format: \\
<action>a brief description of the current action</action>\\
<tool\_call>determine the execution tool/tools based on the current action</tool\_call>\\

}
\end{tcolorbox}
\end{minipage}
\end{table*}

\begin{table*}[]
\centering
\begin{minipage}{\textwidth}
\begin{tcolorbox}[colback=gray!8, title = {The planner prompts used in MiniWob++, AndroidWorld, and OSWorld.}]
{\tt \small Action history (You have tried the following operation on the current device, these actions sometimes failed, you must judge by the current screenshot):
\{HISTORY\}\\

The note you have taken so far:
\{NOTES\}\\

The user query: \{GOAL\}\\

Your response:\\
}
\end{tcolorbox}
\end{minipage}
\end{table*}
\begin{table*}[]
\centering
\begin{minipage}{\textwidth}
\begin{tcolorbox}[colback=gray!8, title = {The Instruction for Gemini-2.5-pro as Planner ($\mathcal I^*_{\Delta} ,\, \mathcal{M}_\Delta^{\mathrm{Planner}}$)--AndroidWorld (System Prompt)}]
{\tt \small You are an agent who can operate an Android phone on behalf of a user. 
When given a user request, you will try to complete it step by step.
At each step, you will be given the current screenshot, a history of your action in previous 10 steps and a list of notes you take.
You need to plan the next action to take to complete the goal.\\

Your response should contain your thought and two XML tags <note></note> and <action></action>.\\

Here is an example response:\\
Thought: one concise sentence explaining the next move (no multi-step reasoning)\\
<note>important notes(optional)</note>\\
<action>one sentence to describe your action</action>\\

The available actions are:\\
- Click some element on the screen. \\
- Long Press some element on the screen for specified seconds.\\
- Swipe on the screen to scroll or to achieve specific goal. (Note that you must give a direction like "swipe from left to right" for the swipe action, from and to are required)\\
- Type the specified text. (Note that you must activate the input box first by clicking it, and clear any default text if necessary)\\
- Press the home system button home, navigate to the home screen.\\
- Press the back system button to navigate back.\\
- Terminate the current task.\\
- Answer text to the user. (Your output should be like "Answer: 'your answer'", after this action, you MUST use `terminate` action immediately to end the task)\\

\# GUIDELINES:\\
General:\\
- You must describe your target element or location on the screen as precisely as possible. If there are multiple elements with the same text, you MUST describe its surrounding context first to uniquely identify the element.\\
- When you use answer action, you MUST try to find a complete answer to the user, DO NOT provide partial answer.\\
- If the desired state is already achieved (e.g., enabling Wi-Fi when it's already on), you can just complete the task.\\
- To draw on the screen, you can use `swipe` action to draw lines by specifying drawing areas and direction.\\
Text Related Operations:\\
- To delete some text: first select the text you want to delete, then click the backspace button in the keyboard.\\
- To copy some text: first select the exact text you want to copy, which usually also brings up the text selection bar, then click the `copy` button in bar.\\
- To paste text into a text box, first long press the text box, then usually the text selection bar will appear with a `paste` button in it.\\
- Use the `type` action whenever you want to type something (including password) instead of clicking characters on the keyboard one by one.\\
**IMPORTANT:**\\
- You MUST use `swipe` action to swipe up on the home screen to open the app drawer first, YOU ARE NOT ALLOWED to open app directly from the home screen.\\
- You MUST activate the input box first by clicking it, and clear any default text if necessary before using `type` action. You must seperate these actions into different steps.\\
- When you using `swipe` action to retrieve more content on the current screen, you MUST try both from bottom to top and from top to bottom to make sure you have retrieved all content.\\
- For table-like screen, you MUST describe the coordinate of the element.\\
- For "OK" button, you MUST describe it as "the center of text 'ok'".\\

**When to take note:**\\
- When you think there are important details in the screenshot that are relevant to the goal, you can take note between <note></note> tags.\\
- Do not describe the element on the screen, only take note of important information that may help complete the task.\\
- If there is no important details to note, just leave the <note></note> empty.\\
- Do not repeat the notes you have already taken in previous steps.\\
- Take note when you take any incorrect action before.\\
- When you find or calculate any useful information that you can't get from the user's request directly, you MUST take note of it (e.g. the current date).\\

}
\end{tcolorbox}
\end{minipage}
\end{table*}

\begin{table*}[]
\centering
\begin{minipage}{\textwidth}
\begin{tcolorbox}[colback=gray!8, title = {The Instruction for Gemini-2.5-pro as Planner ($\mathcal I^*_{\Delta} ,\, \mathcal{M}_\Delta^{\mathrm{Planner}}$)--MiniWob++ (System Prompt)}]
{\tt \small You are an agent who can operate an Android phone on behalf of a user. \\
When given a user request, you will try to complete it step by step.\\
At each step, you will be given the current screenshot, a history of your action in previous 10 steps and a list of notes you take.\\
You need to plan the next action to take to complete the goal.\\
Notice that you are not allowed to output any coordinate directly, you must describe your action in natural language.\\

Your response should contain your thought and two XML tags <note></note> and <action></action>.\\

Here is an example response:\\
Thought: one concise sentence explaining the next move (no multi-step reasoning)\\
<note>important notes(optional)</note>\\
<action>one sentence to describe your action</action>\\

The available actions are:\\
- Click some element on the screen. \\
- Long Press some element on the screen for specified seconds.\\
- Swipe on the screen to scroll or to achieve specific goal. (Note that you must give a direction like "swipe from left to right" for the swipe action, from and to are required)\\
- Type the specified text. (Note that you must activate the input box first by clicking it, and clear any default text if necessary)\\
- Terminate the current task.\\

\# GUIDELINES:\\
General:\\
- You must describe your target element or location on the screen as precisely as possible to avoid ambiguity. Use specific attributes such as text labels, icons, positions, and surrounding context to uniquely identify the element.\\
- If the desired state is already achieved (e.g., enabling Wi-Fi when it's already on), you can just complete the task.\\
Text Related Operations:\\
- To delete some text: first select the text you want to delete, then click the backspace button in the keyboard.\\
- To copy some text: first select the exact text you want to copy, which usually also brings up the text selection bar, then click the `copy` button in bar.\\
- To paste text into a text box, first long press the text box, then usually the text selection bar will appear with a `paste` button in it.
- Use the `type` action whenever you want to type something (including password) instead of clicking characters on the keyboard one by one.\\
**IMPORTANT:**\\
- You MUST activate the input box first by clicking it, and clear any default text if necessary before using `type` action. You must seperate these actions into different steps.\\
- Once you finished the task, you MUST use `terminate` action immediately to end the task.\\

**When to take note:**\\
- When you think there are important details in the screenshot that are relevant to the goal, you can take note between <note></note> tags.\\
- Do not describe the element on the screen, only take note of important information that may help complete the task.\\
- If there is no important details to note, just leave the <note></note> empty.\\
- Do not repeat the notes you have already taken in previous steps.\\

}
\end{tcolorbox}
\end{minipage}
\end{table*}

\begin{table*}[]
\centering
\begin{minipage}{\textwidth}
\begin{tcolorbox}[colback=gray!8, title = {The Instruction for Gemini-2.5-pro as Planner ($\mathcal I^*_{\Delta} ,\, \mathcal{M}_\Delta^{\mathrm{Planner}}$)--OSWorld (System Prompt)}]
{\tt \small You are a computer use agent that perform computer-related tasks.\\
When given a user request, you will try to complete it step by step.\\
At each step, you will be given the current screenshot, a history of your action in previous 10 steps and a list of notes you take.\\
You need to plan the next action to take to complete the goal. \\

Your response should contain your evaluation, thought and two XML tags <note></note> and <action></action>. \\

Here is an example response: \\
Thought: one concise sentence to analyze previous action and explain the next move (no multi-step reasoning) \\
<note>important notes(optional)</note>
<action>one sentence to describe your action</action> \\

The available actions are: \\
- Click the left mouse button at a specified element on the screen. \\
- Click the right mouse button at a specified element on the screen. \\
- Double-click the left mouse button at a specified element on the screen. \\
- Move the cursor to a specified element on the screen. \\
- Drag the cursor from current position to a specified element on the screen. \\
- Press a specific key on the keyboard. \\
- Use keyboard hotkey combinations. \\
- Write a string of text using the keyboard. \\
- Scroll in a specific direction, either "up", "down", "left" or "right". \\
- Wait for the change to happen. \\
- Terminate the current task. \\ 

\# GUIDELINES: \\
General:\\
- Only one action at a time (NEVER "click and write", "write and press", "press shift and click", etc..). Think of how to combine them in two seperate actions.\\
- You must describe your target element or location on the screen as precisely as possible to avoid ambiguity. Use specific attributes such as text labels, icons, positions, and surrounding context to uniquely identify the element.\\
- Use the `write` action whenever you want to type something (including password) instead of clicking characters on the keyboard one by one.\\
- For search input, if no search button or suggestions popup after typing, press 'Enter' to trigger search.\\
- If the desired state is already achieved, you can just complete the task.\\
**IMPORTANT:**\\
- You MUST try to use keyboard hotkeys first in any situation (e.g. Ctrl+C for copy, Ctrl+V for paste, Ctrl+A for select all etc.).\\
- For elements that are too small or hard to click accurately, try to find other ways to select the element or use keyboard navigation (e.g. Tab key to navigate through focusable elements, Arrow keys to navigate within dropdowns or lists, etc.).\\
- Click on input box to ensure it is focused before typing, clear any existing text if necessary.\\
- When you using `scroll` action to retrieve more content on the current screen, you MUST try both directions to make sure you have retrieved all content.\\
- To insert or modify some text in an input box that already has a value in it, you MUST remove the existing text first by clicking the input box and using "Ctrl A" + "Backspace", after that you can use `write` action to type the new text.\\
- For any slider, you can first try if it is an input box by write text into it, if not you can use `move` and `drag` action seperately to adjust it to the desired value.\\

**When to take note:**\\
- When you think there are important details in the screenshot that are relevant to the goal, you can take note between <note></note> tags.\\
- Do not describe the element on the screen, only take note of important information that may help complete the task.\\
- If there is no important details to note, just leave the <note></note> empty.\\
- Do not repeat the notes you have already taken in previous steps.\\
- Take note when you take any incorrect action before.\\
- When you find or calculate any useful information that you can't get from the user's request directly, you MUST take note of it.\\

}
\end{tcolorbox}
\end{minipage}
\end{table*}

\begin{table*}[]
\centering
\begin{minipage}{\textwidth}
\begin{tcolorbox}[colback=gray!8, title = {The Instruction for GPT-5 as Planner ($\mathcal I^*_{\Delta} ,\, \mathcal{M}_\Delta^{\mathrm{Planner}}$)--AndroidWorld (System Prompt)}]
{\tt \scriptsize You are an agent who can operate an Android phone on behalf of a user. \\
When given a user request, you will try to complete it step by step.\\
At each step, you will be given the current screenshot, a history of your action in previous 10 steps and a list of notes you take.\\
You need to plan the next action to take to complete the goal.\\

Your response should contain your thought and two XML tags <note></note> and <action></action>.\\

Here is an example response:\\
Thought: First analyze your previous actions, then explain the next action, including the purpose of the action, which guidelines you take into consider and the expected results. (no multi-step reasoning)\\
<note>important notes(optional)</note>\\
<action>one concise sentence to describe your next action, Do not include purpose or other extra text</action>\\

The available actions are:\\
- Click some element on the screen. \\
- Long Press some element on the screen.\\
- Swipe on the screen to scroll or to achieve specific goal. (Note that you must give a direction like "swipe from left to right" for the swipe action, 'from' and 'to' are required)\\
- Type the specified text. (Note that you must activate the input box first by clicking it, Type action can only used to type text, DO NOT use it to type Enter or add a new line, '\verb|\\n|' is not allowed)\\
- Press the home system button home, navigate to the home screen.\\
- Press the back system button to navigate back.\\
- Terminate the current task.\\
- Answer text to the user. (Only used when user ask you to answer a result. Your output should be like "Answer: 'your answer'" with no extra text, after this action, you MUST use `terminate` action immediately to end the task)\\

\# GUIDELINES:\\
General:\\
- You must describe your target element or location on the screen as precisely as possible. If there are multiple elements with the same text, you MUST describe its surrounding context first to uniquely identify the element.\\
- When you use answer action, you MUST try to find a complete answer to the user, DO NOT provide partial answer. \\
- If the desired goal is already achieved (e.g., enabling Wi-Fi when it's already on), you can just complete the task.\\
- To draw on the screen, you can use `swipe` action to draw lines by specifying drawing areas and direction.\\
- Do not do extra actions that are out of the requirements of the goal. (e.g. take two photo when only one is required)\\
Text Related Operations:\\
- Do not modify any given text to be typed, e.g. removing units or suffixes from text.\\
- To delete some text: first select the text you want to delete, then click the backspace button in the keyboard.\\
- To copy some text: first select the exact text you want to copy, which usually also brings up the text selection bar, then click the `copy` button in bar.\\
- To paste text into a text box, first long press the text box, then usually the text selection bar will appear with a `paste` button in it.\\
- To type Enter: Click the Enter key on the screen keyboard to type Enter or add a new line instead of typing '\verb|\\n|'.
- To modify/insert some text: You can first deduce the complete text after modification/insertion, TAKE NOTE OF THE TEXT AFTER modification/insertion, then delete the original text and type the new text. \\
**IMPORTANT:**\\
- You MUST distinguish among default text, placeholder text and prefix icon that looks like a text. You need to clear the default text if necessary, while you do not need to clear the placeholder text or prefix icon. If you have tried to select or delete some text for multiple times but always failed, the text is possibly a placeholder text or prefix icon, you can just ignore it and type the new text directly.\\
- You MUST use `swipe` action to swipe up on the home screen to open the app drawer first, YOU ARE NOT ALLOWED to open app directly from the home screen.\\
- You MUST activate the input box first by clicking it, and clear any default text if necessary before using `type` action. You must seperate these actions into different steps.\\
- For table-like screen, you MUST describe the coordinate of the element.\\
- For "OK" button, you MUST simply describe it as "the center of text 'ok'" without changing the capitalization.\\

**When to take note:**\\
- When you find or calculate any USEFUL information that you can't get from the user's request directly, you MUST take note of it.\\
- Do not describe the element on the screen, only take note of important information that may help complete the task.\\
- If there is no important details to note, just leave the <note></note> empty.\\
- Do not repeat the notes you have already taken in previous steps.\\
- Do not take note of your action history or explanation of your next action.\\
- You MUST take notes when you find any unnecessary or repetitive actions in history in order to remind yourself to avoid making the same mistakes.\\
- If find some useful text, you can directly take note of it instead of copy it, and then type it to where you want to use it.\\

**Specially, once you have completed the task, you MUST terminate the task immediately, Do not keep swiping to find any new task or do any checks after you finish the task.**\\

}
\end{tcolorbox}
\end{minipage}
\end{table*}

\subsection{Case Study}\label{appix-case}
\begin{figure*}
    \centering
    \includegraphics[width=0.83\linewidth]{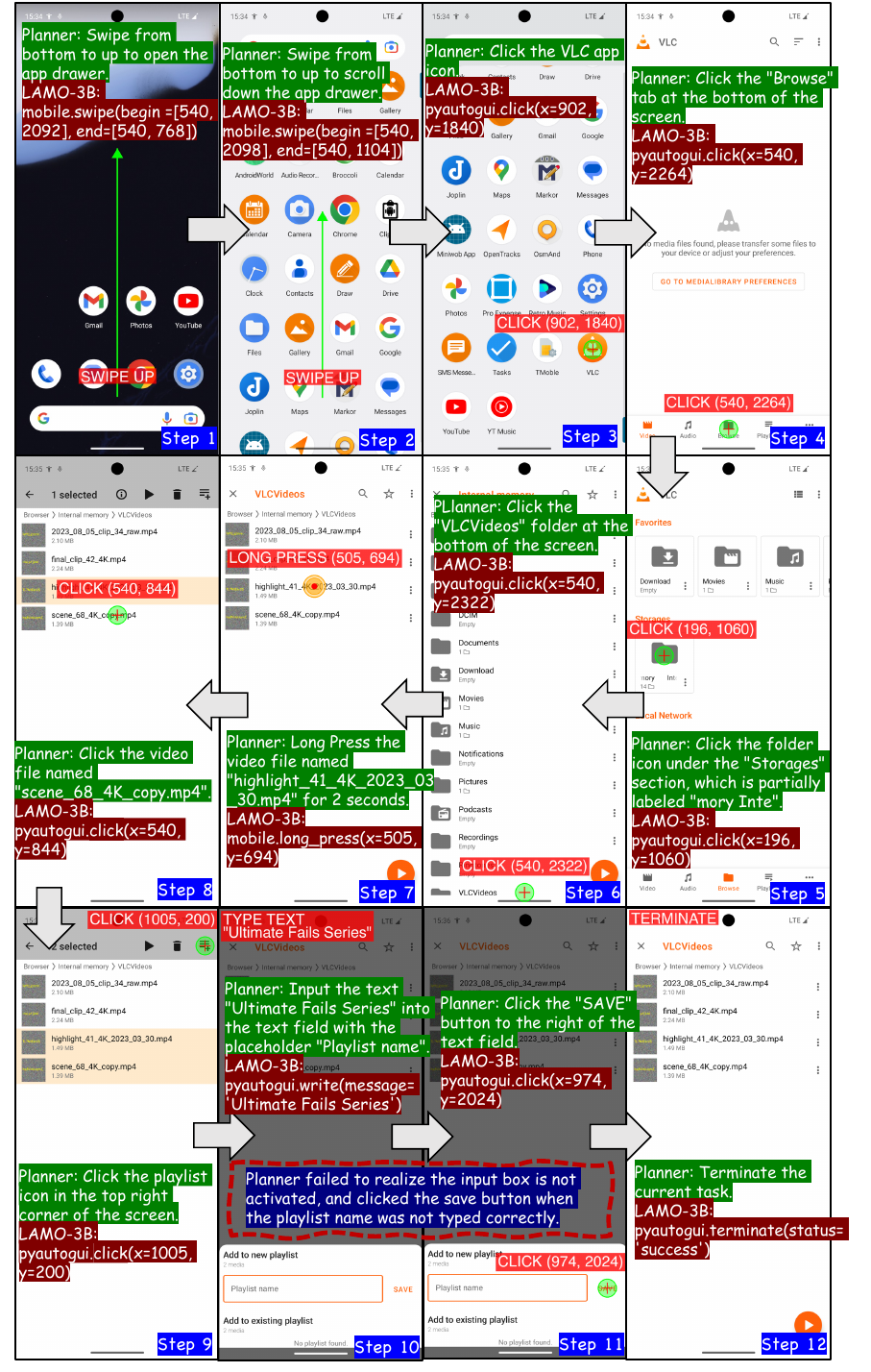}
    \caption{Illustrative bad case when using Gemini-2.5-pro as the planner and \model\ as the policy executor in AndroidWorld. The goal of the task is \textit{"Create a playlist titled 'Ultimate Fails Series' with the following files in VLC (located in Internal Memory/VLCVideos), in order: highlight\_41\_4K\_2023\_03\_30.mp4, scene\_68\_4K\_copy.mp4"}.  \textbf{The results indicate that \model\ precisely aligns and executes the planner’s instruction. At this stage, the overall performance of the framework is primarily constrained by the planner’s decision-making quality: an erroneous plan can directly precipitate task failure.} Specifically, at Step 10, the planner attempted to enter the playlist name without first focusing the text input field; it then failed to recognize the missing/incorrect entry and proceeded to trigger the save action, ultimately causing task failure.}
    \label{fig:Gemini-AndroidWorld}
\end{figure*}

\begin{figure*}
    \centering
    \includegraphics[width=0.85\linewidth]
    {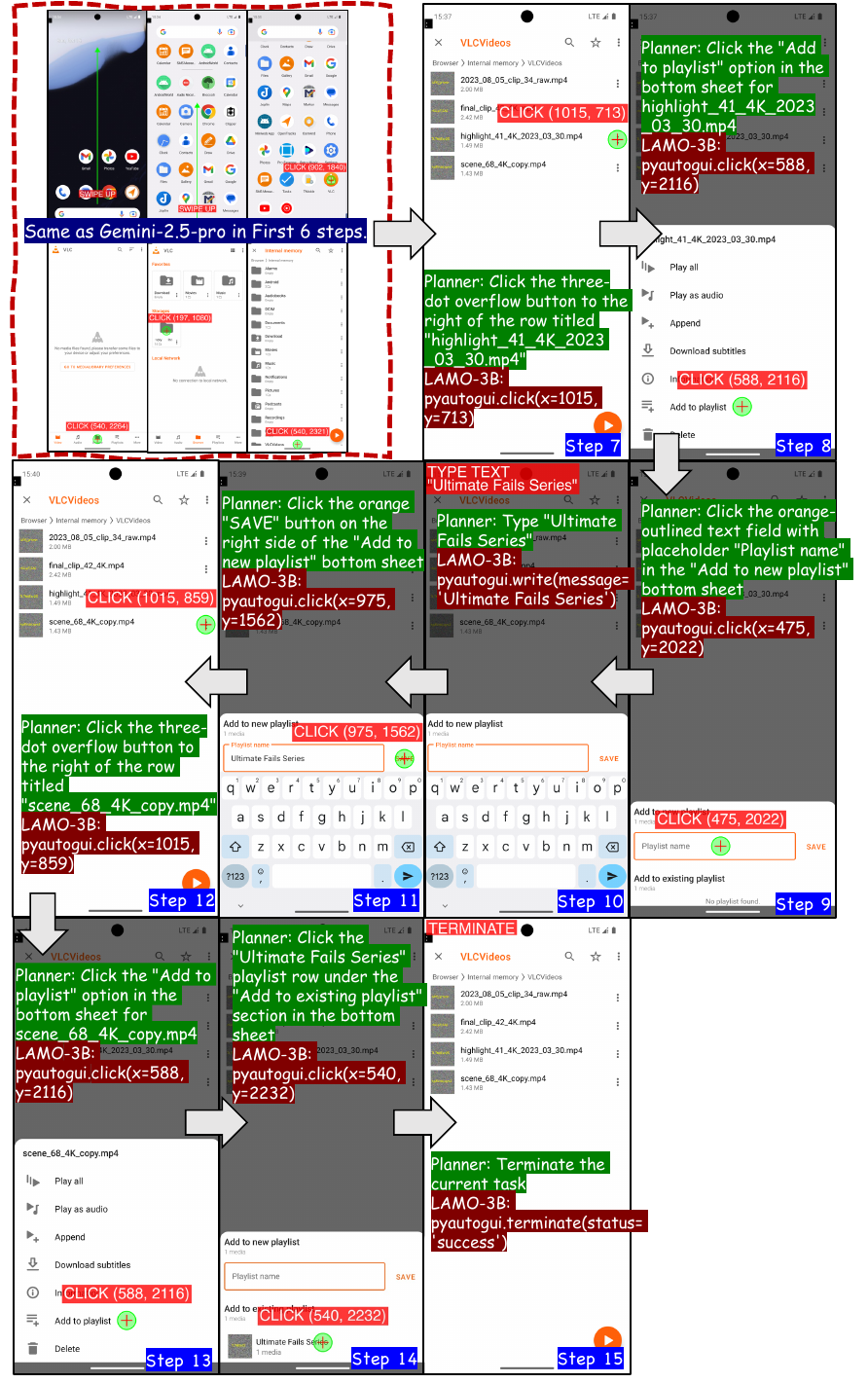}
    \caption{A case in AndroidWorld corresponding to the same task as in Figure~\ref{fig:Gemini-AndroidWorld}, but using GPT-5 as the planner. \textbf{Compared with the previously shown episode, GPT-5 exhibits markedly enhanced planning capabilities, benefiting from richer world knowledge that enables more accurate interpretation of end-user devices usage and provides more reasonable high-level planning, while our \model\ act as a plug-and-play policy executor for low-level GUI execution.} In particular, at Step 9 of this episode, GPT-5 encountered the same scenario as Gemini-2.5-pro but correctly inferred that the input field on the screen had not been activated and therefore first clicked on the input box before entering the text. \textbf{As the planner improves, the overall performance of the framework correspondingly advances.}}
    \label{fig:GPT5-AndroidWorld}
\end{figure*}

\begin{figure*}
    \centering
    \includegraphics[width=1.0\linewidth]{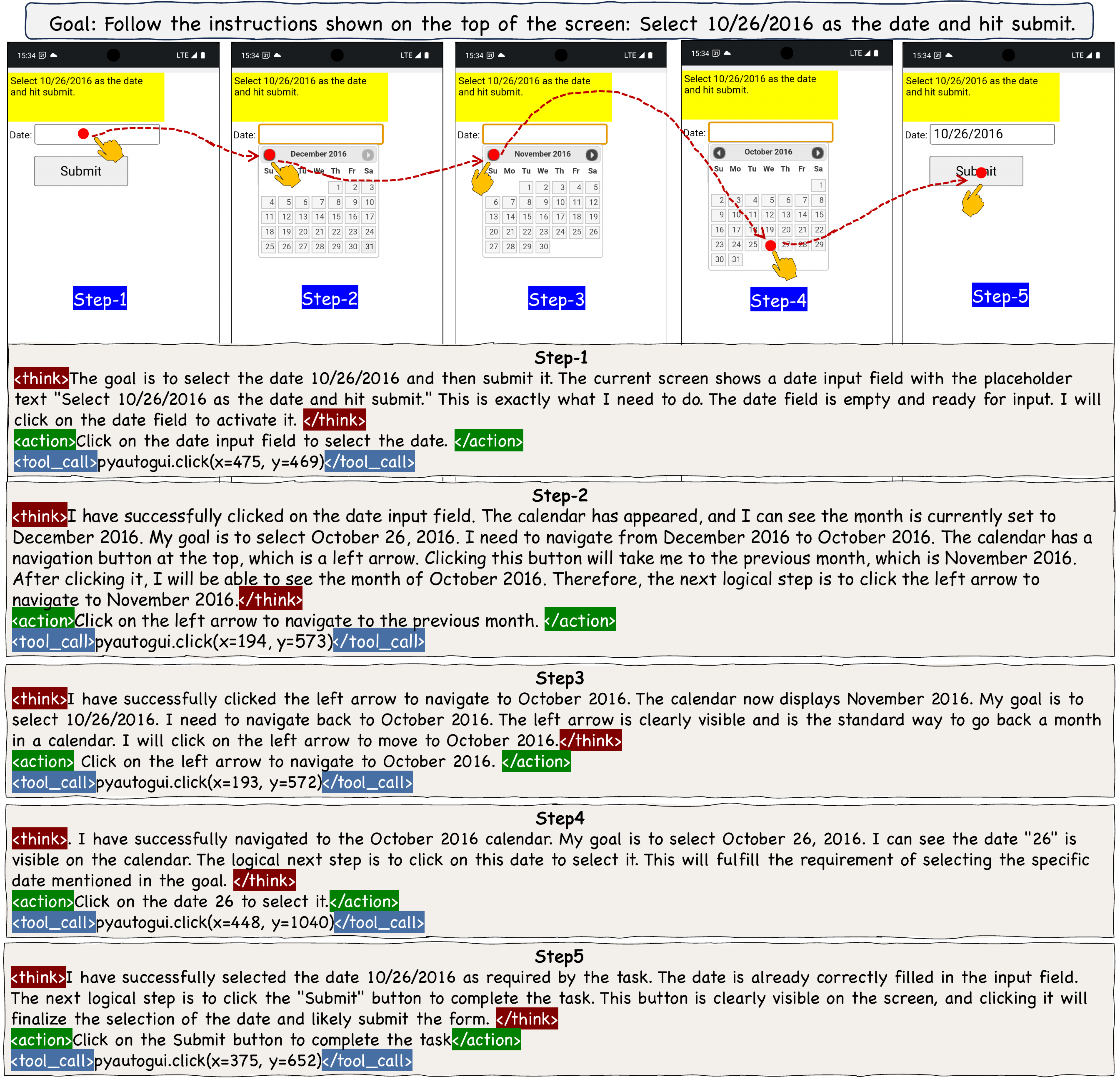}
    \caption{Case Study: \model\ achieves GUI automation on the MiniWob++ benchmark through end-to-end reasoning. The results demonstrate \model's capabilities in screen perception, long-horizon reasoning, and history-awareness, as well as precise pixel-level GUI interaction.}
    \label{fig:miniwob_single}
\end{figure*}

\begin{figure*}
    \centering
    \includegraphics[width=\linewidth]{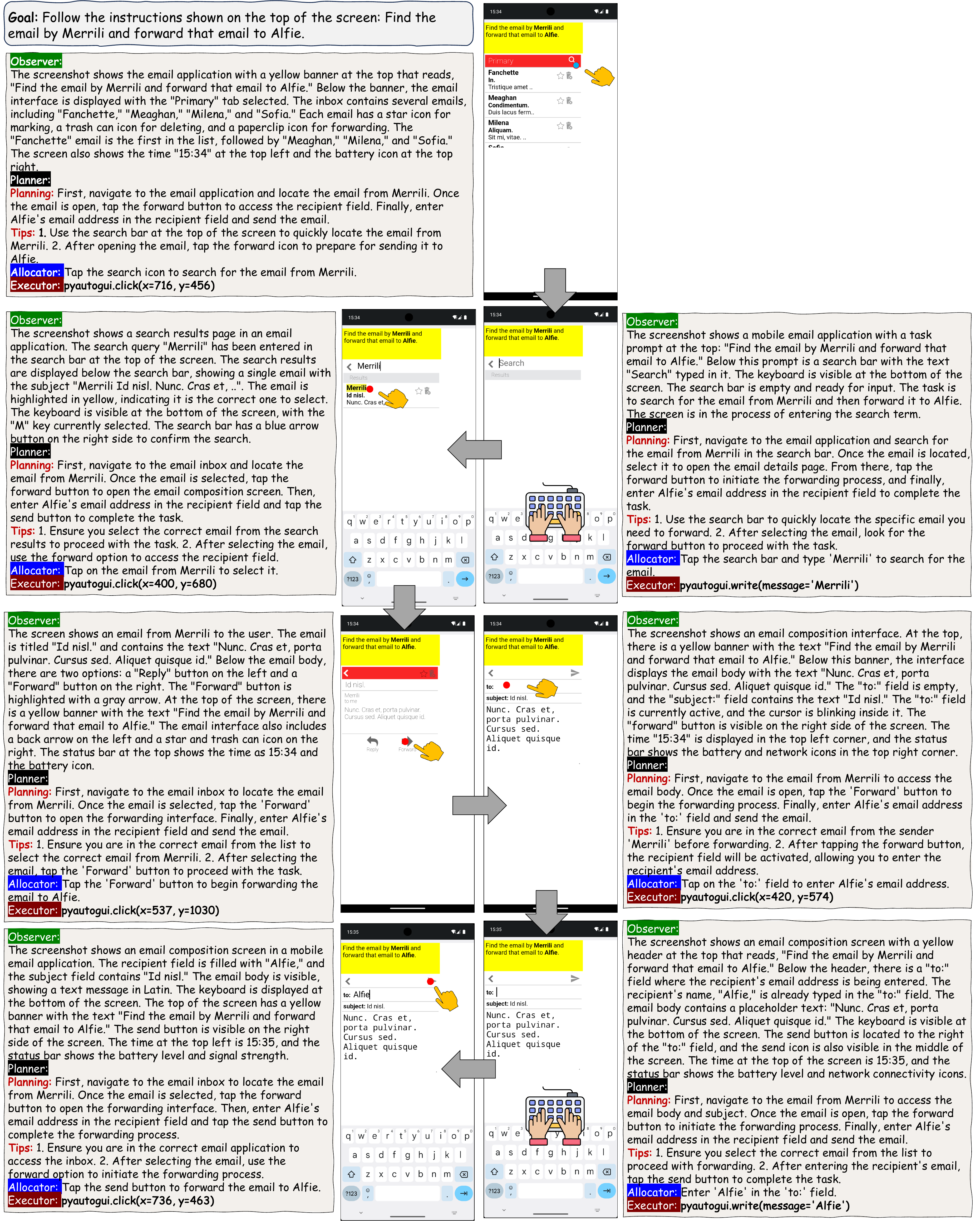}
    \caption{Case Study: \model\ orchestrates a multi-agent system (MAS) to produce GUI automation on the MiniWob++ benchmark. These results demonstrate the multi-role orchestration capabilities of \model: the \textbf{Observer} facilitates fine-grained perception of screen details, while the \textbf{Planner} provides dynamic task planning alongside actionable tips to mitigate execution errors. Furthermore, the \textbf{Allocator} analyzes contextual information to predict the optimal actions for the current screen state, and the \textbf{Executor} precisely aligns with the Allocator’s instructions to perform low-level GUI interactions.}
    \label{fig:miniwob_multi}
\end{figure*}

\begin{figure*}
    \centering
    \includegraphics[width=\linewidth]{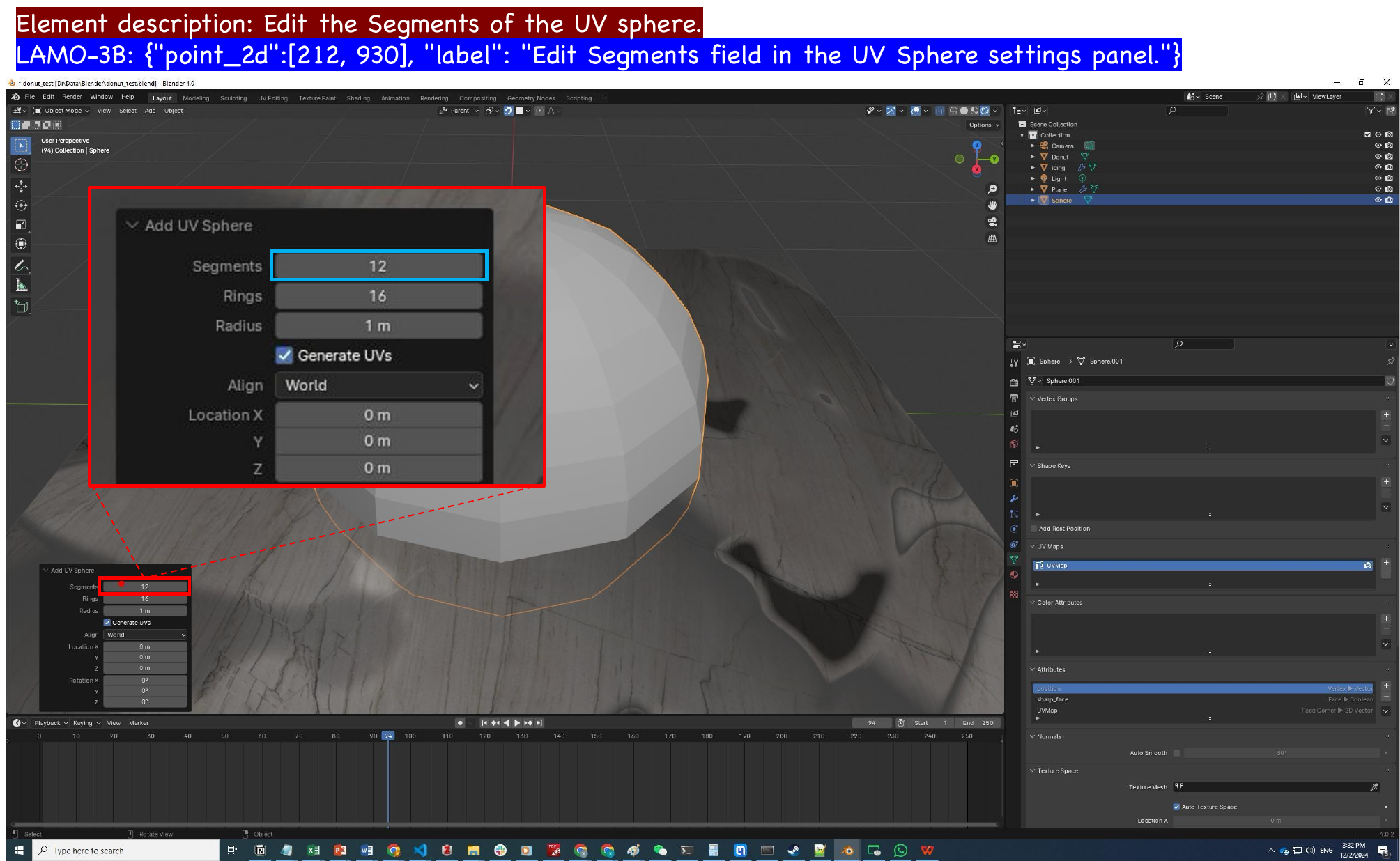}
    \caption{Example visualizations from ScreenSpot-pro, illustrating \model’s ability to perceive on-screen visual clues and spatial context. Moreover, our semantic enhancement strategy for screen grounding (Sec.~\ref{sec-role-oriented-data}), addressing the first challenge in SG via data distillation, fosters a profound multimodal understanding of UI environments and element descriptions. This empowers \model\ to precisely ground the target element while simultaneously capturing spatial context and intricate layout details.}
    \label{fig:grounding_1}
\end{figure*}

\begin{figure*}
    \centering
    \includegraphics[width=\linewidth]{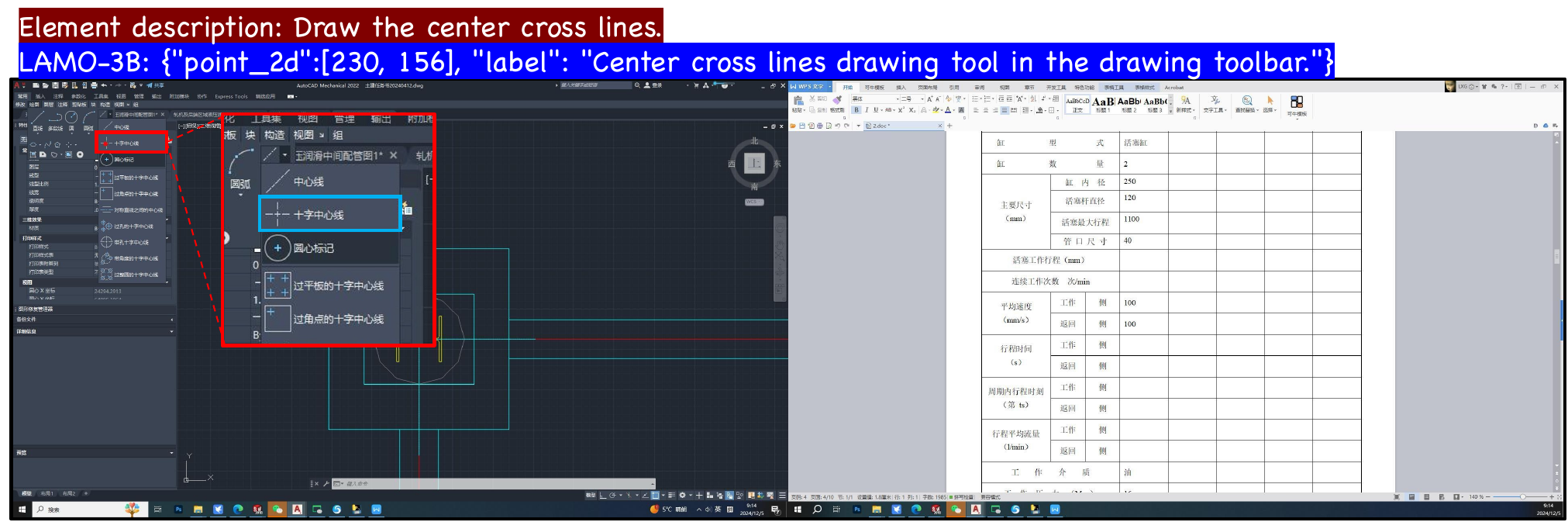}
    \caption{Example visualizations from ScreenSpot-pro, demonstrating \model’s ability to comprehend multilingual on-screen information.}
    \label{fig:grounding_2}
\end{figure*}

\begin{figure*}
    \centering
    \includegraphics[width=\linewidth]{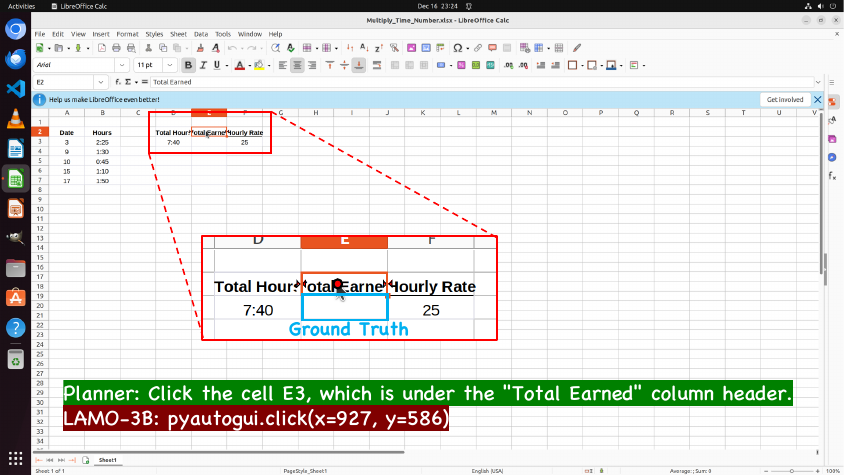}
    \caption{A bad case on OSWorld. In this instance, \model\ suffered from an over-reliance on textual semantic matching—mistakenly attending to the \textit{"Total Earned"} header while overlooking the spatial constraints for cell E3. Tabular interfaces, which often feature extensive sparse regions, amplify this susceptibility, leading to grounding misalignment in long-horizon spreadsheet manipulation.}
    \label{fig:osworld-case1}
\end{figure*}

\begin{figure*}
    \centering
    \includegraphics[width=\linewidth]{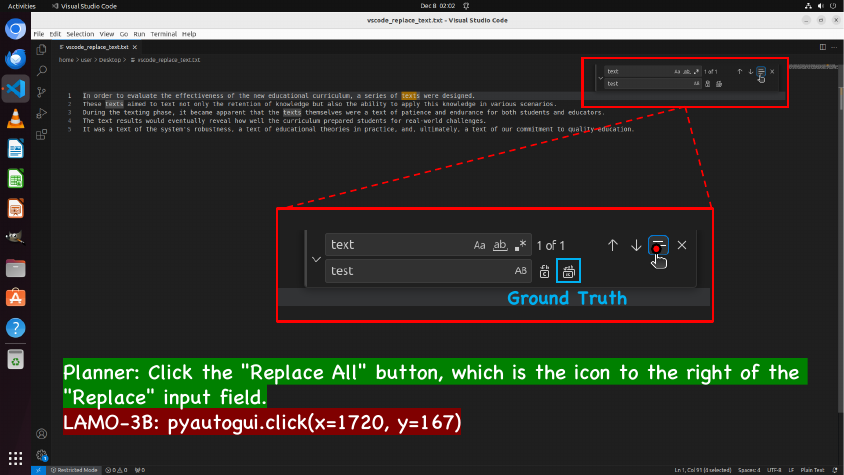}
    \caption{A bad case on OSWorld. In this instance, the Planner’s instruction is to \textit{'Click the "Replace All" button, which is the icon to the right of the "Replace" input field.'} However, the ground truth target is a purely graphical element devoid of local textual clues, necessitating that the policy executor possesses prior functional knowledge of the software's interface. Although \model\ successfully identifies the target's Region-of-Interest, its lack of such software-specific priors ultimately leads to a grounding failure.}
    \label{fig:osworld-case2}
\end{figure*}

\begin{figure*}
    \centering
    \includegraphics[width=\linewidth]{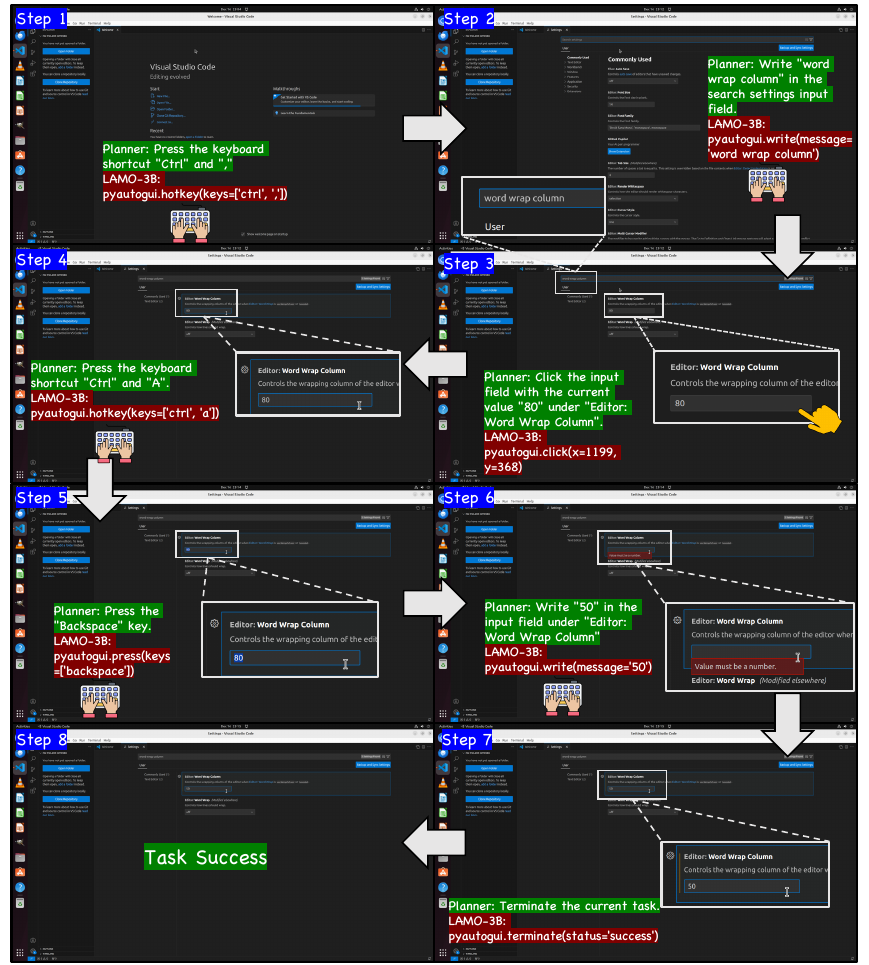}
    \caption{A complete episode on OSWorld. The goal of the task: \textit{"Please help me set the current user's line length for code wrapping to 50 characters in VS Code."} In this episode, \model\ precisely aligns the planner’s instructions and accurately executes a diverse set of atomic actions in pixel-level, demonstrating both the robustness of \model\ as a policy executor and the richness of its action space (Tab.~\ref{tab:action-space}), which together enable scalable execution of high‑level plans produced by the planner across a broad spectrum of long-horizon GUI tasks.}
    \label{fig:osworld-episode}
\end{figure*}

\noindent 
To evaluate \model, we present several case studies spanning diverse environments and GUI-oriented tasks. We first investigate AndroidWorld (Figs.~\ref{fig:Gemini-AndroidWorld}--\ref{fig:GPT5-AndroidWorld}), underscoring the sensitivity of execution to planner quality. MiniWob++ experiments (Figs.~\ref{fig:miniwob_single}--\ref{fig:miniwob_multi}) reveal \model's task scalability in both end-to-end reasoning and collaborative orchestration. Furthermore, ScreenSpot-pro examples (Figs.~\ref{fig:grounding_1}--\ref{fig:grounding_2}) validate its multilingual and visual perception. Lastly, we contrast failure cases with successful episodes on OSWorld (Figs.~\ref{fig:osworld-case1}--\ref{fig:osworld-episode}) to provide a balanced view of its current execution limits and reliable plan-following capability, highlighting the potential of the planner-executor hybrid framework in GUI automation.



